\documentclass[11pt]{article}

\usepackage[final]{acl}

\usepackage{times}
\usepackage{latexsym}

\usepackage[T1]{fontenc}
\usepackage[utf8]{inputenc}

\usepackage{microtype}

\usepackage{inconsolata}

\usepackage{graphicx}

\usepackage{hyperref}
\usepackage{url}
\usepackage[utf8]{inputenc} % allow utf-8 input
\usepackage[T1]{fontenc}    % use 8-bit T1 fonts
\usepackage{hyperref}       % hyperlinks
\usepackage{url}            % simple URL typesetting
\usepackage{booktabs}       % professional-quality tables
\usepackage{amsfonts}       % blackboard math symbols
\usepackage{nicefrac}       % compact symbols for 1/2, etc.
\usepackage{microtype}      % microtypography
\usepackage{xcolor}         % colors

\usepackage{bookmark}
\usepackage{graphicx}
\usepackage{CJKutf8}
\usepackage{xpinyin}
\xpinyinsetup{ratio=0.6}

\usepackage{algorithmic}
\usepackage{algorithm}
\usepackage{amsmath}
\usepackage{amsfonts}
\usepackage{booktabs}
\usepackage{array}
\usepackage{tabularx}
\usepackage{diagbox}
\usepackage{arydshln}
\usepackage{multirow}
\usepackage{float}
\usepackage{natbib}
\usepackage{wrapfig}
\usepackage{amssymb}
\usepackage{tcolorbox}
\usepackage{listings}

\newtheorem{theorem}{Theorem}

\newtheorem{proof}{Proof}[section]

\newcommand{\name}[0]{{{C$^2$TU}}}

\title{Breaking the Cloak! Unveiling Chinese Cloaked Toxicity with Homophone Graph and Toxic Lexicon}

\author{%
  Xuchen Ma\textsuperscript{1},
  Jianxiang Yu\textsuperscript{1}, 
  Wenming Shao\textsuperscript{2}, 
  Bo Pang\textsuperscript{2}, 
  Xiang Li\textsuperscript{1}\thanks{Corresponding Author}
  \\
  \textsuperscript{1}School of Data Science and Engineering, East China Normal University\\
  \textsuperscript{2}Shanghai EastWonder Info-tech Co., Ltd.\\
  \tt{\{xuchenma, jianxiangyu\}@stu.ecnu.edu.cn} \\
  \tt{\{simon, pangbo\}@wdit.com.cn} \\
  \tt{xiangli@dase.ecnu.edu.cn}\\
}

\begin{document}
\maketitle

\begin{CJK*}{UTF8}{gbsn}
\begin{abstract}
Social media platforms have experienced a significant rise in toxic content, including abusive language and discriminatory remarks, presenting growing challenges for content moderation. Some users evade censorship by deliberately disguising toxic words through homophonic cloak, which necessitates the task of unveiling cloaked toxicity. Existing methods are mostly designed for English texts, while Chinese cloaked toxicity unveiling has not been solved yet. To tackle the issue, we propose \name, a novel training-free and prompt-free method for Chinese cloaked toxic content unveiling. It first employs substring matching to identify candidate toxic words based on Chinese homophone graph and toxic lexicon. Then it filters those candidates that are non-toxic and corrects cloaks to be their corresponding toxicities. Specifically, we develop two model variants for filtering, which are based on BERT and LLMs, respectively. For LLMs, we address the auto-regressive limitation in computing word occurrence probability and utilize the full semantic contexts of a text sequence to reveal cloaked toxic words. Extensive experiments demonstrate that \name\ can achieve superior performance on two Chinese toxic datasets. In particular, our method outperforms the best competitor by up to 71\% on the F1 score and 35\% on accuracy, respectively. 
Our code and data are available at
\url{https://github.com/XDxc-cuber/C2TU-Chinese-cloaked-toxicity-unveiling}.
% \url{https://github.com/XDxc-cuber/C2TU}.
\end{abstract}

% \begin{center}
{
\textcolor{red}{\textbf{Disclaimer}:\textit{ The paper contains content that may be profane, vulgar, or offensive.}}
}

% \end{center}

\section{Introduction}

With the exponential growth of user-generated contents, online social media platforms have become an inevitable communication tool with massive users~\cite{omipidan2024rise, Briukhanova2021social, bala2014social}. 
Although the rapid information dissemination has clear benefits, a significant amount of toxic contents have emerged on social platforms over the past decade, such as abuse, discrimination, and cyberbullying~\cite{1, recuero2024platformization, kim2021distorting, hanscom2024toxicity}. Social networks are thus facing increasingly severe challenges in governing toxicity. 

Toxicity detection has recently attracted extensive attention~\cite{anjum2024hate, lima2024toxic, schmidhuber2024llm}, but most of the proposed methods ~\cite{16,17,14,zhao2024enhancing} can only output binary predictions rather than recover the original toxic expressions hidden in cloaked text.
Further,
to evade censorship, some users on social platforms
intentionally disguise toxic words by 
% intentionally 
replacing parts or all of the words with homophonic characters or emojis~\cite{6, 17, kirk-etal-2022-hatemoji}.
For example,
% in Chinese social media, some users 
replace ``操''~(\pinyin{cao4}, means ``f*ck'') with ``草''~(\pinyin{cao3}, means ``grass''), or ``垃圾''~(\pinyin{la1ji1}, means ``rubbish'') with ``辣鸡''~(\pinyin{la4ji1}, means ``spicy chicken'') in Chinese.
These deliberate cloaks significantly degrade the effectiveness of existing toxicity detection methods \cite{6} and also 
% allowing them to post toxic content. 
necessitate the task of \emph{unveiling cloaked toxicity}, i.e., restoring cloaked toxic words to their original proto-toxic words.
While there have been some methods~\cite{13,10} proposed to solve the task,
most of them are specially designed for English texts,
% Further, existing methods for cloaked toxicity correction~\cite{13,10} are mostly designed for English, 
which cannot be directly used for Chinese due to the different characteristics of the two languages.
Further,
% for Chinese cloaked toxic content,
although a recent work \cite{6}
% although [xxx] 
points out the existence of Chinese cloak toxic content,
they fail to offer a solution.
Therefore, 
to bridge the gap,
% beyond detection,
a research question naturally arises:
\textit{Can we develop a model to unveil Chinese cloaked toxic contents?}

% Meanwhile,
% To solve the problem,
We notice that 
{Chinese Spelling Correction (CSC)}, which aims to correct misspelled Chinese characters, 
shares some similarities with our task.
% is a task that 
% which shares some similarities with our task.
% the task of Chinese cloaked toxic content correction.
% methods seem to provide a workable way to correct cloaked toxic comments, as involving correcting misspelled Chinese characters.
However, the two tasks are not entirely equivalent. 
While CSC deals with unintentional user typos, misspellings in toxic content are often deliberate. 
Moreover, the corpora used in CSC tasks typically consist of more standardized language, which differs significantly in distribution from the toxicity found on the internet. 
Therefore, although models like SCOPE~\cite{23} and Simple-CSC~\cite{12} have demonstrated strong performance on CSC tasks, directly applying them to this restoration task may not yield satisfactory 
% similarly effective
results (see Section~\ref{sec:exp}).

% \xc{Therefore, although existing methods such as SCOPE~\cite{23} and Simple-CSC~\cite{12} have demonstrated strong performance on CSC tasks, directly applying them to the correction of Chinese cloaked toxic contents may not yield satisfied 
% similarly effective
% results.}

\begin{figure*}[t]
\centering
  \includegraphics[width=1\linewidth]{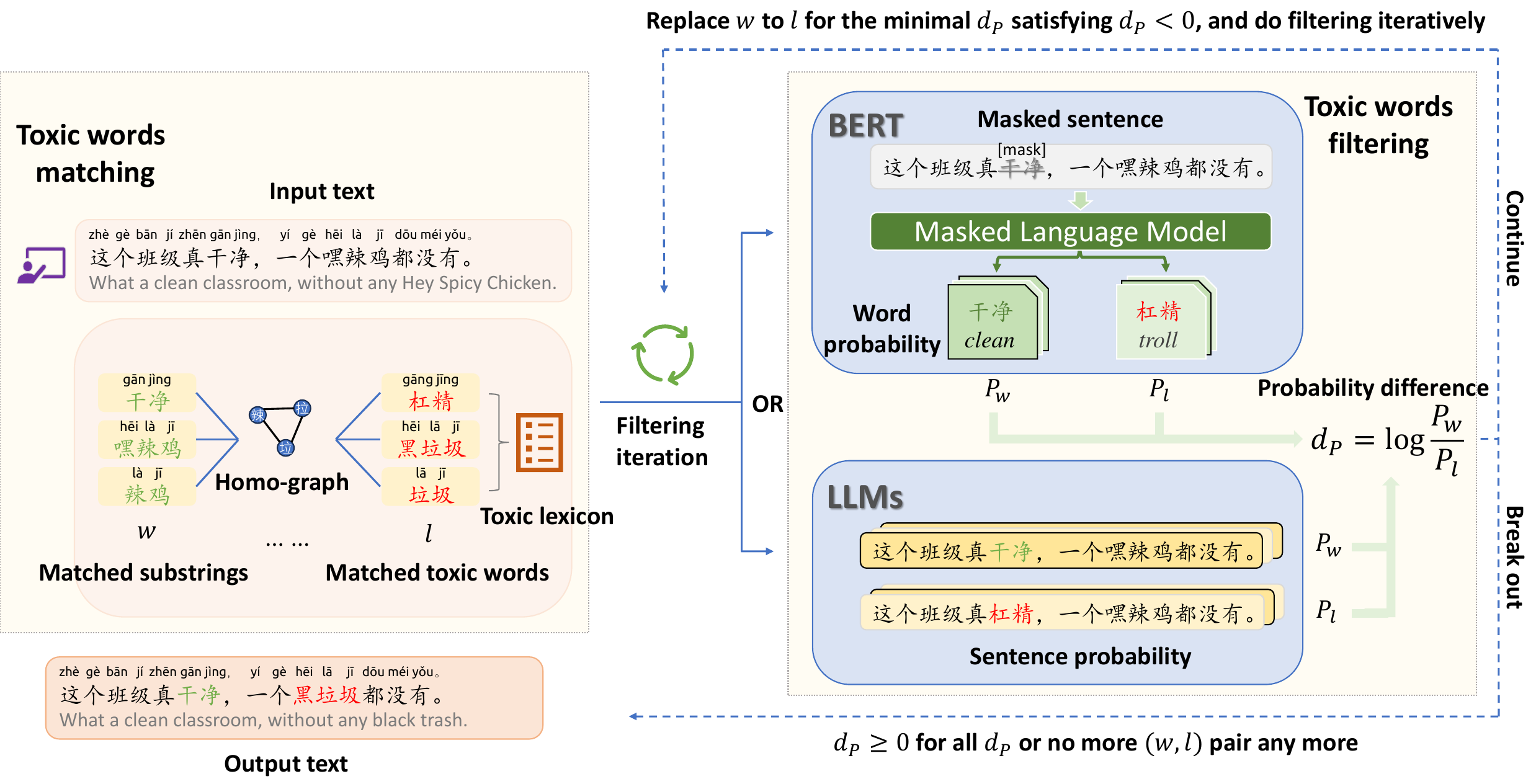}
  \caption {The main workflow of \name\ method. \name\ consists of two key stages: matching and filtering. In the matching stage, we identify candidate toxic terms in the input text by leveraging a homophone graph and a toxic lexicon. In the filtering stage, we use a language model (BERT model or LLM) to compute the probability gap for each \((w, l)\) pair and iteratively replace the pair with the most significant gap. Once the iteration terminates, the unveiled sentence is returned.}
  \label{fig:main_fig}
\end{figure*}

% To address the problem, 
In this paper,
we study \underline{C}hinese \underline{C}loaked \underline{T}oxicity
\underline{U}nveiling as a sequence restoration task,
and propose the \name\ model (see Figure~\ref{fig:main_fig}).
% based on homophone relationships and Chinese toxic lexicon.
% to correct Chinese cloaked toxic comments based on BERT models or LLMs. 
Specifically,
we utilize a homophone graph and a toxic lexicon to identify potential toxic words within the input text, thereby transforming the problem into a candidate toxic words filtering task. 
This task aligns well with BERT model's pretraining objective of masked language modeling. 
% Motivated by this, 
Hence, we first employ BERT model to compute the probability difference between tokens derived from raw and toxic words at specific positions in a sentence.
If toxic tokens have larger probabilities,
we then unveil the cloak.
% to correct cloaked toxicity.
% deciding whether toxic words should occur.
% achieving promising results.
% With the advent of LLMs, represented by GPT-3~\cite{brown2020language} and LLaMA~\cite{19}, their strong generalization and language understanding capabilities have led to remarkable performance across a wide range of natural language processing~(NLP) tasks. 
After that,
we further explore using LLMs to address the filtering task. 
Similar to the BERT-based approach, we attempt to calculate word occurrence probabilities to decide whether a replacement should occur. 
% However, we observe that this naive application of LLMs results in suboptimal performance, and \li{in some cases even amplifies noise}.
% Inspired by BERT model’s masked language model architecture, we hypothesize that the problem lies in the auto-regressive nature of LLMs, which only conditions on the left-side context and thus lacks access to right-side semantics. 
To better exploit full sentence context,
we reformulate the filtering score as a sentence-level likelihood difference.
% converting.
Since LLMs are well-suited for computing sentence-level likelihoods, this formulation allows us to incorporate the full semantic context of a sentence. 
% Ablation study shows that this sentence-probability-based method surpasses the BERT-based approach.
% In addition, 
% we also design a multi-round iterative filtering strategy to progressively denoise toxic text. 
% By iteratively correcting the most probable words pair in each round, the semantic fluency of the sentence is significantly improved, enabling the language model to make more accurate judgments in subsequent iterations. 
% Ablation study confirms the effectiveness of our iterative strategy.
Finally,
we summarize the main contributions of our paper as:

\begin{itemize}
\item 
We formulate Chinese cloaked toxicity unveiling as a sequence restoration task and propose a training-free and prompt-free method \name\  
for Chinese cloaked toxicity unveiling.
Different from most existing methods for English and from generic toxicity detection,
to our best knowledge,
we are the first to solve the problem under Chinese.
% to address the intentional substitution of Chinese toxic words on social platforms. 
% Our method leverages toxic lexicons and Chinese character homophonic relationship graphs to assist language model's correction.
% \item[$\bullet$] 
% We present two methods 
% \name-BERT and \name-LLM to filter 
% candidate toxic words.
% In particular,
% we
% mathematically transform

% to leverage the full semantic context of a sentence for calculating the appearance probability of a word,
\item
We propose BERT- and LLM-based filtering modules for candidate toxic words.
In particular,
the LLM variant uses a sentence-level likelihood formulation so it can exploit the full semantic context during filtering.

% For BERT, it leverages the full semantic contexts of a sentence to predict the probability of a word in the sentence.

% We utilize BERT model's mask filling capability and LLM's language modeling capability on toxic words filtering task. We find that the probability difference of different tokens appearing in a sentence can be transformed into the form of a probability difference between two constructed sentences, which is more suitable for computation by LLMs.
% \item[$\bullet$] 
\item
We conduct extensive experiments to evaluate the model performance on two Chinese toxic datasets.
The results show that our methods are more competitive than other baselines w.r.t. both F1 score and accuracy metrics. In particular,
our method outperforms the best competitor by up to 71\% on the F1 score and 35\% on accuracy.

% Our method significantly outperforms all baseline approaches on the majority of evaluation metrics across two mainstream Chinese toxic datasets. Specifically, it achieves sentence-level correction F1 scores of 70\% on the \texttt{ToxicloakCN} dataset and 88\% on the \texttt{COLDataset}.
\end{itemize}

\section{Related Work}
% In this section, we summarize related work on toxicity correction and Chinese spelling correction, respectively.\xc{删掉这段}

\subsection{Cloaked Toxicity Unveiling}

While
extensive researches have been conducted on toxic content detection~\cite{3,7,8,9,16,17,14,15},
% However, 
% current 
most of them 
% prove ineffective when 
are not specially designed for
% users deliberately 
cloaked toxic words~\cite{6}. 
% especially meeting keyword-based detection methods~\cite{20}, which are widely used in social platforms. 
Further, the reveal of deliberately cloaked toxic words emerges as a critical research challenge worthy of attention.

Some recent efforts have been made to address the issue by handling intentional obfuscation through character scrambling and misspellings in English. For example, some works~\cite{13} leverage contextual information to correct intentionally misspelled words, while others~\cite{10} focus on addressing word order disruptions by employing LLMs for robust comprehension of noisy text. However,
these methods are designed for English.
Unlike English—a word-based language where obfuscation typically alters intra-word characters—Chinese operates at the character level but cloaks content at the word level.
This fundamental linguistic disparity prevents direct application of existing English-based methods to Chinese.
However,
% To our best knowledge,
research on unveiling Chinese cloaked toxic content remains largely unexplored.
% Yet, our survey reveals no recent work addressing Chinese cloaked toxic contents correction, creating an urgent need for tailored solutions.

Recent Chinese toxicity studies have also expanded the problem setting with multimodal benchmarks, span-level target-aware extraction, and homophonic-perturbation detection~\cite{yang-etal-2025-exploring-multimodal,bai-etal-2025-state,wu-etal-2025-enhancing-chinese}. These works strengthen detection and extraction resources, but they still stop short of restoring cloaked spans to their original toxic expressions.

% The primary application scenario for toxic content detection is social media platform, where users are the primary sources of toxic information. Currently, most platforms rely on keyword-based detection methods~\cite{20}.  Efforts have been made to address this issue by handling intentional obfuscation through character scrambling and misspellings in English. For instance, \citet{13} leveraged contextual information to correct intentionally misspelled words, while \citet{10} focused on addressing word order disruptions by employing LLMs for robust comprehension of noisy text.

% This phenomenon is also prevalent in the context of Chinese social media platforms. \citet{17} curated a benchmark dataset that specifically includes toxic content with substitutions of homophonic words. Similarly, \citet{6} considered both homophone and emoji substitutions as deliberate cloak approaches, constructing a noise-augmented dataset \texttt{ToxicloakCN} based on \texttt{ToxiCN}~\cite{15}. While these studies have recognized the challenge in the context of Chinese and provided relevant datasets~(or not), effectively mitigating this issue in Chinese text remains an open research challenge.

\subsection{Chinese Spelling Correction}

Our task shares similarities with the CSC task, which aims to identify and correct Chinese spelling errors based on contextual, phonetic and graphemic information. 
% The CSC task originates from spelling detection and grammatical error identification in English.
Some existing methods are based on small language models (e.g., BERT)~\cite{21,22,23}.
For example, SCOPE~\cite{23} utilize phonetic knowledge of Chinese characters in conjunction with BERT for spelling correction.
With the advent of LLMs, 
there are also LLM-based models~\cite{11,12,li2024c}.
A representative model 
Simple-CSC~\cite{12} is a training-free and prompt-free method, treating LLMs purely as language models to perform spelling correction at the token level.

% For language models, \citet{21} propose the Soft-Masked BERT approach, which employs two separate network models for error detection and correction. \citet{22} incorporate phonetic and graphemic similarities between Chinese characters, leveraging a graph convolutional network~\cite{GCN} and BERT to correct spelling errors. Similarly, \citet{23} utilize additional phonetic knowledge of Chinese characters in conjunction with BERT for spelling correction.

% For LLMs, \citet{11} introduced radicals and pronunciation information as prompts, combined with few-shot learning, to enable large models to correct erroneous text directly. \citet{12} proposed a training-free and prompt-free method, treating LLMs purely as language models to perform spelling correction at the token level. \citet{li2024c} observed that, before and after tokenizing, the Chinese characters and the tokens do not align one-to-one, which led to poor performance of the large language model on character-level spelling correction tasks. They retrained the large language model using a vocabulary that maps Chinese characters to tokens in a one-to-one manner.

However, the CSC task is not well aligned with our task, as the substitution of toxic words is often intentional, and toxic texts themselves are abnormal texts that deviate from regular language usage. Therefore, we pay attention to Chinese cloaked toxicity unveiling in this paper.
% we propose \name\  to address the task of correcting misspellings in toxic Chinese content.

\section{Methodology}
\label{methodology}

\subsection{Graph and Lexicon Construction}

\subsubsection{Chinese Homo-graph}
\label{sc:homo}

% In this section, we present the construction of our Chinese homophone relationship graph, which is named homophone graph.
Given a Chinese toxic speech dataset,
we first construct a homophone graph $\mathcal{G} = (\mathcal{N}, \mathcal{E})$ to capture the likely cloak between tokens, 
where each node in $\mathcal{N}$ represents a Chinese character, and each edge in $\mathcal{E}$ indicates the phonetically similar relation.
Specifically,
we extract all Chinese characters from the dataset as nodes, then use the open-source library pypinyin\footnote{\url{https://pypi.org/project/pypinyin/}} to obtain their pinyin regardless of tones. 
Then characters with identical pinyin are connected. 
% Note that 
Each node also has a self-loop, as each character is considered to have a homophonic relation with itself.
We further consider polyphonic characters and phonetically similar pronunciations based on regional dialects.
% First, 
% we use the open-source library pypinyin\footnote{\url{https://pypi.org/project/pypinyin/}} to construct an undirected, unweighted Chinese homophone graph $\mathcal{G} = (\mathcal{N}, \mathcal{E})$, where each node represents a Chinese character, and each edge indicates a phonetic similarity relationship between two characters. 

% Given a Chinese toxic text dataset,
% we first extract all Chinese characters from the dataset as graph nodes, then use pypinyin to obtain their pinyin regardless of tones. 
% Then characters with identical pinyin are connected. 

% Further,

\noindent
\textbf{Polyphonic Characters:} Some Chinese characters have multiple pronunciations. If a character has at least one pronunciation matching another character's, an edge is linked between them.

\noindent
\textbf{Dialectal Relationships:} 
There also exist dialectal phonetic confusions in Chinese, where users may replace characters based on their dialect pronunciations. 
Common confusions include:

% \begin{itemize}
% \item Retroflex and non-retroflex sounds\cite{dong2006automatic}, such as ``山''~(\pinyin{shan1}, means ``mountain'') and ``三''~(\pinyin{san1}, means ``three'').

% \item Front and back nasal sounds\cite{chen2011perceptual}, such as ``应''~(\pinyin{ying1}, means ``should'') and ``因''~(\pinyin{yin1}, means ``reason'').

% \item Initial consonants ``n'' and ``l''\cite{zhou2019erp, koffi2019acoustic}, such as ``男''~(\pinyin{nan2}, means ``male'') and ``蓝''~(\pinyin{lan2}, means ``blue'').
% \end{itemize}

\noindent $\bullet$ Retroflex and non-retroflex sounds\cite{dong2006automatic}, such as ``山''~(\pinyin{shan1}, means ``mountain'') and ``三''~(\pinyin{san1}, means ``three'').

\noindent $\bullet$ Front and back nasal sounds\cite{chen2011perceptual}, such as ``应''~(\pinyin{ying1}, means ``should'') and ``因''~(\pinyin{yin1}, means ``reason'').

\noindent $\bullet$ Initial consonants ``n'' and ``l''\cite{zhou2019erp, koffi2019acoustic}, such as ``男''~(\pinyin{nan2}, means ``male'') and ``蓝''~(\pinyin{lan2}, means ``blue'').

Considering these confusions, we introduce five types of additional relations: \textit{``n'' $\leftrightarrow$ ``l'', ``zh'' $\leftrightarrow$ ``z'', ``ch'' $\leftrightarrow$ ``c'', ``sh'' $\leftrightarrow$ ``s'' and ``*ng'' $\leftrightarrow$ ``*n''}. 
For example, ``男''~(\pinyin{nan2}) and ``蓝''~(\pinyin{lan2}) share an \textit{``n'' $\leftrightarrow$ ``l''} confusion, so an edge is added between node ``男'' and node ``蓝'', although their pinyin representations are not entirely identical.

\subsubsection{Toxic Lexicon}

We also utilize a toxic lexicon to introduce external knowledge about toxicity. 
The lexicon is a set of toxic words denoted as $\mathcal{L} = \{l_1, \cdots, l_m\}$.
While we notice that some public datasets~(e.g., \texttt{ToxiCN}~\cite{15}) have their own toxic lexicon released,
they also contain many toxic words with homophone cloak. As these cloaks are frequently and commonly used on social platforms, they gradually become substitutes of their corresponding toxic contents. 
We take ``垃圾''~(\pinyin{la1ji1}, means ``rubbish'') and ``辣鸡''~(\pinyin{la4ji1}, means ``spicy chicken'') as an example. 
Here, the proto-toxic word is ``垃圾'', while ``辣鸡'' is a cloaked substitution commonly used in social media, even more frequently than the proto-toxic word.
% Each dataset employs its specific toxic lexicon.
% Based on the original toxic lexicon from existing works, 
% we conduct denoising and deduplication processes. 
% The original toxic lexicon of its dataset contains a large number of toxic words with homophone cloak. 
These cloaks should also be removed instead of remaining in the vocabulary as toxic knowledge.  
Therefore,
to keep a simple and clean toxic lexicon, 
% ensure that the toxic lexicon is clean, 
we only retain proto-toxic words.

\subsection{Toxic Word Matching}
\label{sc:WordMatching}

% Based on the Chinese homophone graph and toxic lexicon, we designed a toxic word matching algorithm. 
% \xc{Given a text sequence $X$, we employ a homophone graph $ \mathcal{G} $ and a toxic word lexicon $ \mathcal{L} $ with size $S$ to match all possible candidate toxic words.
% Formally, for each toxic word 
% $ l^{(i)} = \{ l^{(i)}_1, l^{(i)}_2, \dots, l^{(i)}_N \} \in \mathcal{L} $, 
% where $ N $ denotes the length of the word, we match it against all substrings 
% $ w^{(j)} = \{ w^{(j)}_1, w^{(j)}_2, \dots, w^{(j)}_N \} $ of the same length in $ X $. 
% If and only if for all positions $ \forall k \in \{1, 2, \dots, N\} $, the following condition holds:
% $
% \mathcal{G}.\text{HasEdge}(l^{(i)}_k, w^{(j)}_k) = 1, 
% $
% then the substring $ w^{(j)} $ is considered as a \textbf{candidate toxic word}. 
% Here,
% $\mathcal{G}.\text{HasEdge}(\cdot, \cdot) = 1$ indicates that two nodes are linked in the homophone graph $\mathcal{G}$.}

Given a text sequence $X$, we enumerate substring $w$ in $X$ and match each $w$ against toxic words $l\in \mathcal{L}$ using the homophone graph $\mathcal{G}$. 
% \li{Specifically, 
% we execute the suspected toxic word matching algorithm\ref{alg:stwma} to detect potentially cloaked toxic words}.
Formally, a substring $w$ of length $N$ is considered as a candidate toxic word if and only if $\exists l\in\mathcal L$, ${\rm{len}}(w) = {\rm{len}}(l)$ and $\forall k\in\{1,2,\cdots,N\}$, 
% we have
% \begin{equation}
${\rm{\mathcal G.HasEdge}}(w_{k},l_{k})=1$,
% \end{equation}
where $w_{k}$ is $k$-th Chinese character in $w$, and so does $l_{k}$.
% We constrain that $w_i$ and $l_j$ are of the same length.
% The lengths of $w_i$ and $l_j$ are equal, i.e., ${\rm{len}}(w_i) = {\rm{len}}(l_j) = N$, where 
Here, ${\rm{len}}(\cdot)$ is the length of the input string. 
% the function that yields string's length. 
The function ${\rm{\mathcal G.HasEdge}}(\cdot, \cdot) = 1$ indicates that two nodes are linked in $\mathcal{G}$.

The above matching method could have a small \textit{false negative rate}, i.e.,
any homophonic substitution will be unveiled 
% any toxic word from the lexicon that has been substituted via homophonic substitution will be detected 
given a comprehensive homophone graph $\mathcal{G}$ and lexicon $\mathcal{L}$. 
However, it may also yield a high \textit{false positive rate} by incorrectly matching non-toxic word as toxic word. For example, ``干净''~(\pinyin{g}{\pinyin{an1}} \pinyin{jing4}, means ``clean'') will be matched to ``杠精''~(\pinyin{g}{\pinyin{ang4}} \pinyin{jing1}, means ``troll''), even though ``干净'' is typically a non-toxic word. 
False positives are especially severe for single-character toxic words, as all homophones of the single toxic character~(e.g., ``操''~(\pinyin{cao4}, means ``f*ck'')) would be matched.
% This is clearly unreasonable.
The pseudocode of the matching algorithm is summarized in Alg.~\ref{alg:stwma},
which outputs 
$M$ candidate toxic words pairs, denoted as $W_p=\{(w^{(i)},l^{(i)})\}$ for $i=1,\cdots, M$.
Here, $l^{(i)}$ is a probable toxic unveiling of $w^{(i)}$.

% we execute the suspected toxic word matching algorithm\ref{alg:stwma} to detect potentially cloaked toxic words

% To address this issue, we propose filtering methods based on BERT and open-source LLMs to refine detected potential toxic words.

% meaning they share a similar homophonic relationship. 
% If there exists a pair $(w_i, l_j)$ satisfying the above condition, the substring $w_i$ in $X$ is deemed a potential toxic word.

\begin{figure*}[t]
\centering
  \includegraphics[width=1\linewidth]{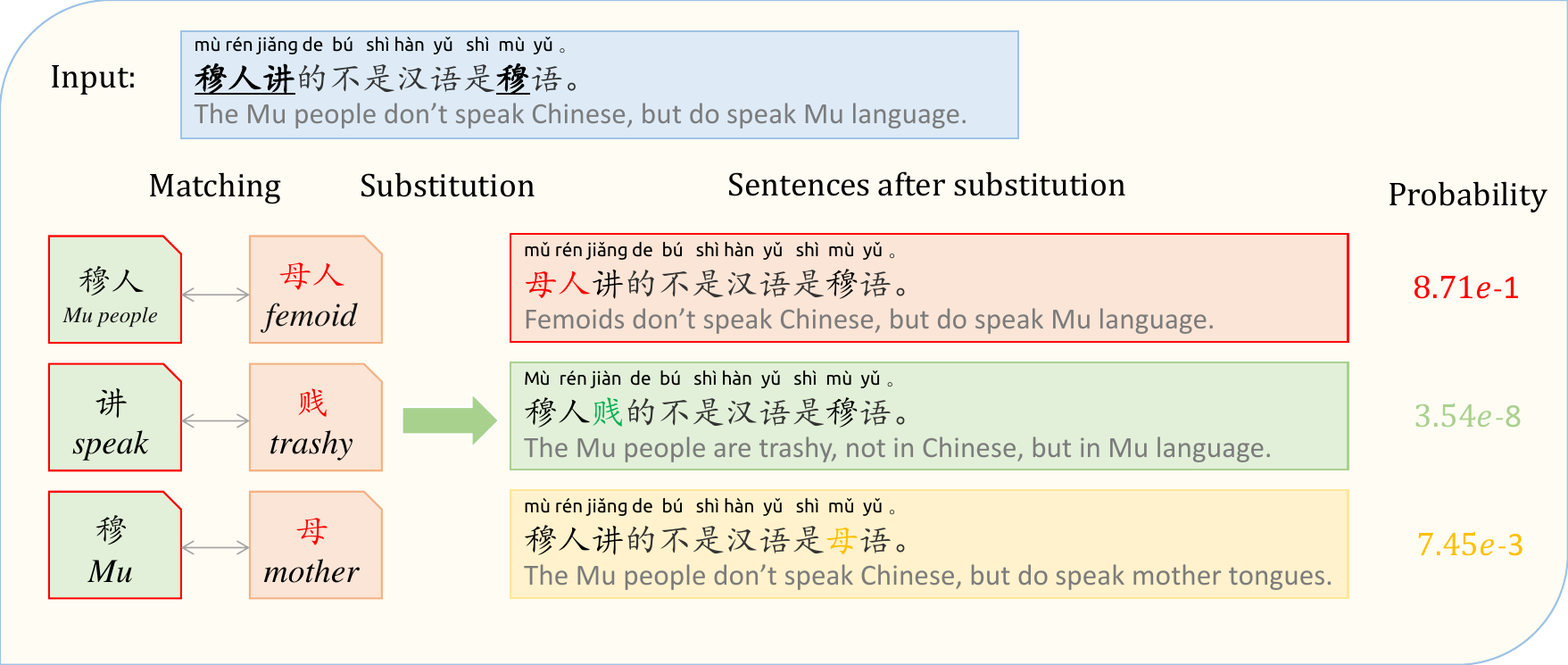}
  \caption {Three candidate toxic words pairs are matched in the input text. Each pair is individually substituted to generate three new sentences. It is evident that the pair [``讲'', ``贱''] is incorrect, as the resulting sentence is highly unnatural in meaning, leading to a significantly smaller probability. In contrast, the substitutions of [``穆人'', ``母人''] and [``穆'', ``母''] are correct, and their corresponding sentences yield higher probabilities.}
  \label{fig:example}
\end{figure*}

\subsection{Filtering Candidate Toxic Words}
\label{sc:FilterWords}

% For a given user comment text $X$, the toxic word matching algorithm\ref{alg:stwma} identifies $M$ potentially toxic words pairs, denoted as $W_p=\{(w_i,l_i)\}$ for $i=1,2,\cdots,M$. Here, $w_i$ and $l_i$ represent the potentially toxic word and the proto-toxic word corresponding to the $i$-th group, respectively.

To further filter out the incorrect matches (see Figure \ref{fig:example}) in $W_p$,
we leverage the full semantics of the text sequence $X$.
% The algorithm for filtering potentially toxic words needs to incorporate the semantic information of $X$ in order to filter out unreasonable replacements~(such as the previously mentioned ``干净'' and ``杠精''). This results in a subset $W_p'\subseteq W_p$.
% Firstly, we define $X_{\text{pre}},X_{\text{tail}}$.
Formally, for each $(w,l)$ pair,
we define 
\( X_{\text{pre}} \) and \( X_{\text{tail}} \) as the prefix sequence and suffix sequence of \( w \) in \( X \), respectively. 
% the original position of $w_i$ in $X$ is denoted as $p_i$.
% Then we have:
% such that:
% \begin{equation}
% \begin{aligned}
% X_{\text{pre}}&=[x_1,\cdots,x_{p_i-1}],\\
% X_{\text{tail}}&=[x_{p_i+\text{len}(w_i)},\cdots,x_T]
% \end{aligned}
% \end{equation}
% That is, \( X_{\text{pre}} \) and \( X_{\text{tail}} \) represent the prefix and suffix of \( X \) after removing \( w_i \). 
% The fundamental goal is to
% Our goal is to compare
Given \( X_{\text{pre}} \) and \( X_{\text{tail}} \),
we next calculate the probabilities \(  P(w | X_{\text{pre}}, X_{\text{tail}}) \) and \(  P(l | X_{\text{pre}}, X_{\text{tail}}) \),
shorted as $P_{w}$ and $P_{l}$, respectively.
% In each iteration, for all \( i \), we calculate \(  P(w_i | X_{\text{pre}}, X_{\text{tail}}) \) and \(  P(l_i | X_{\text{pre}}, X_{\text{tail}}) \). 
For all pairs \( (w, l) \) that satisfy 
% \(  P(w_i | X_{\text{pre}}, X_{\text{tail}}) <  P(l_i | X_{\text{pre}}, X_{\text{tail}}) \) 
{\(P_{w}<P_{l}\)}, 
% $P_{w_i} < P_{l_i}$,
we select the one
% \( k \) 
that yields the largest difference, denoted as \( (\tilde{w}, \tilde{l}) \).
% the most significant difference between these two probabilities. 
Formally,
the probability difference is calculated by:{\begin{equation}
% \nonumber
% \begin{aligned}
\text{ProbDiff}(P_{w}, P_{l}) = \log P_{w} - \log P_{l} = \log\frac{P_{w}}{P_{l}}.
% \end{aligned}
\end{equation}}
% \noindent

After that, in the given text sequence $X$, 
we replace \( \tilde{w} \) with \( \tilde{l} \), remove \( (\tilde{w}, \tilde{l}) \) from $W_p$ and then repeat the previous process until 
$W_p$ is empty or all the remaining pairs in $W_p$ 
have 
% all the pairs in $W_p$ is traversed.
% there are no remaining potentially toxic words, or the computed probabilities for all potentially toxic words indicate that 
% \(  P(w_i | X_{\text{pre}}, X_{\text{tail}}) \ge  P(l_i | X_{\text{pre}}, X_{\text{tail}}) \) 
{\(P_{w}\ge P_{l}\)}.
Note that in each iteration, 
we need to 
% The reason for recalculating 
recalculate the probabilities for each pair in $W_p$
because 
% after the replacement is that, 
once \( \tilde{w} \) is replaced, 
\( X_{\text{pre}} \) and \( X_{\text{tail}} \) for remaining pairs change.

\subsection{Computing Probability Difference}
% Our objective is to compute \(  P(x | X_{\text{pre}}, X_{\text{tail}}) \) using language models.
In this section, we introduce two methods for computing \(  P(x | X_{\text{pre}}, X_{\text{tail}}) \), where \(x\) can be either \(w\) or \(l\), with both BERT and LLMs. 
% \xc{with both BERT model and LLM.}

% \xc{能否不定义x，因为LLM方法里还是用的w和l，BERT方法里可以只描述w，然后说l是同理的。}

\subsubsection{BERT-based Method}

% We found that BERT's~\cite{2} mask-filling capability is formally well-suited for \(  P(x | X_{\text{pre}}, X_{\text{tail}}) \).

% BERT is a NLP model based on the Transformer architecture~\cite{4}, achieving excellent performance in a variety of NLP tasks. 
During the pre-training phase, BERT utilizes an unsupervised Masked Language Modeling~(MLM) task, whose goal is to predict tokens that are randomly masked.
% where tokens in the text are randomly masked and the model is tasked with predicting the masked tokens. 
Hence,
% In this way,
% the mask-filling capability of BERT is
% This also makes it
BERT is well-suited for computing \(  P(x | X_{\text{pre}}, X_{\text{tail}}) \).
% Formally, for an input sequence
% \begin{equation}

% \xc{$x_i$的记号是否不准确，前面$w_i$表示第$i$个$w$，这里$x_i$又表示token，是不是可以改成$x^{(i)}$.}

Given a text sequence $X = [X_{\text{pre}},x, X_{\text{tail}}]$,
for the $i$-th token $x_i$ in $x$, 
we first mask it with the [\texttt{mask}] token and 
derive a text sequence $X^{\text{mask}}_{i}=[X_{\text{pre}}, \cdots, x_{i-1},[\texttt{mask}], x_{i+1}, \cdots, X_{\text{tail}}]$.
Then we can output the probability for the masked token \( P_{\text{BERT}}(x_i|X_{i}^{\text{mask}})\). For a toxic word $x$ that includes $N$ tokens,
we sequentially predict each of these tokens and combine them by: 
\begin{equation}
% \nonumber
% \small
\begin{aligned}
 P_{\text{BERT}}(x|X_\text{pre},X_\text{tail})=\ &\prod_{i=1}^{N} P_{\text{BERT}}(x_i|X^{\text{mask}}_{i}).
\end{aligned}
\label{eq:bertx}
\end{equation}

% Considering the difference in probability magnitudes 
Due to the varying word lengths, 
we adjust Equation~\ref{eq:bertx} to  calculate the geometric mean of all $N$ probabilities. 
% based on the word lengths.
% This, in terms of logarithms, is equivalent to computing the average value. 
Finally,
the log probability to predict $x$ is formally given by: 
% Therefore, the final word probability based on the BERT model is given as:
% \li{
% \begin{equation}
% \begin{aligned}
%     \text{ProbDiff}(p^{\text{BERT}}_{w},& p^{\text{BERT}}_{l}) =\\
%     & \frac{1}{N}(\log p_w^{\text{BERT}}-\log p_l^{\text{BERT}})
% \end{aligned}
% \end{equation}
% }
\begin{equation}
\begin{aligned}
\log P_x 
% &=
    % \log\bar{ P}_{\text{BERT}}(x|X_\text{pre},X_\text{tail}) \\
    &= \log\sqrt[N]{{ P}_{\text{BERT}}(x|X_\text{pre},X_\text{tail})}\\
    &=\frac{1}{N}\sum_{i=1}^N\log P_{\text{BERT}}(x_i|X^{\text{mask}}_{i})
\end{aligned}
\end{equation}

\subsubsection{LLMs-based Method}
\label{sec:prellm}
We next explore to use LLMs to filter candidate toxic words.
% Motivated by the outstanding language capabilities of LLMs, we attempted to utilize LLMs to filter potential toxic words. 
% Current LLMs are based on auto-regressive architecture. The generation process is completed step by step, with each step relying only on the previously generated content. 
% More formally, s
% Suppose a sequence \( (x_1, x_2, \cdots, x_n) \). The auto-regressive language model learns the joint probability distribution: 
% \begin{equation}
% P(t_1, t_2, \cdots, t_n) = \prod_{i=1}^{n} P(t_i | t_1, t_2, \cdots, t_{i-1})
% \end{equation}
% Therefore, auto-regressive language models are inherently unidirectional, meaning they can only infer future tokens based on preceding tokens and cannot utilize future context to predict preceding tokens.
Due to the auto-regressive nature,
mainstream LLMs
% a straightforward way to 
% they 
compute the occurrence probability of $x$ based on the pre-context $X_{\text{pre}}$ by:
% \(  P(x | X_{\text{pre}}, X_{\text{tail}}) \) by:
% using the auto-regressive probability distribution is given as follows:
\begin{equation}
% \nonumber
\begin{aligned}
    P_{\text{LLM}}(x|X_\text{pre}&)=\prod_{i=1}^{N}P_{\text{LLM}}(x_i|X_\text{pre},x_1,\cdots,x_{i-1}).
\end{aligned}
\label{eq:llmauto}
\end{equation}

% This method can partially estimate the probability of \( x \) appearing at the middle of $X_\text{pre},X_\text{tail}$,
In this way,
% However, 
only pre-context is used while the post-context is totally ignored.
% its contextual reference is limited to \( X_{\text{pre}} \) and lacking information from \( X_{\text{tail}} \).
To address the issue, 
instead of directly computing the probability for a word,
we score each candidate by comparing the likelihood of the original sentence and the sentence after replacement.
This yields a sentence-level form that can use the full context, as shown below.
Formally, we show:
% for comparing the probabilities of different words' appearance within auto-regressive language models. Our ultimate goal is to compare two probabilities' difference, that is, to compute \(  P(w | X_{\text{pre}}, X_{\text{tail}}) -  P(l | X_{\text{pre}}, X_{\text{tail}}) \).

% We can prove that, under the auto-regressive architecture of large language models (LLMs), the difference in the probabilities of two words appearing at the same position in a sentence is equivalent to the difference in the probabilities of the two corresponding complete sentences where each word is substituted, respectively.

\begin{theorem}
\label{theo}
Given a text sequence $X = [X_{\text{pre}},w, X_{\text{tail}}]$
and a toxic word $l$ with ${\mathrm{len}}(w) $= ${\mathrm{len}}(l)$ = $N$,
let $X'$ denote the new sequence with $w$ replaced by $l$. Then, the following relation holds:
\begin{equation}
    {\mathrm{ProbDiff}}(P_w,P_l) = \log P_{\mathrm{LLM}}(X)-\log P_{\mathrm{LLM}}(X'),
\end{equation}
% \begin{equation}
% \begin{aligned}
%     {\mathrm{ProbDiff}}&(p_w,p_l)\\
%     =\ &\log P_{\mathrm{LLM}}(X)-\log P_{\mathrm{LLM}}(X')
% \end{aligned}
% \end{equation}
where 
$ P_{\mathrm{LLM}}(X)$ and $ P_{\mathrm{LLM}}(X')$
denote the probability of $X$ and $X'$ output by LLM, respectively. Both the probabilities can be calculated in the auto-regressive manner as Equation~\ref{eq:llmauto}.
\end{theorem}

% \begin{table*}[t!]
% \centering
% \caption{The main results of all baseline models and our method w.r.t. the F1 score metric, where D-F means the F1 score of detection and C-F means the F1 score of correction. We highlight the best score in bold and underline the runner-up one.}
% \label{tab:main-result-f1}
% \input{tables/main_tab_f1}
% \end{table*}

% \begin{table*}[t!]
% \centering
% \caption{The main results of all baseline models and our method w.r.t. the accuracy metric, where D-A denotes detection accuracy and C-A denotes correction accuracy.}
% \label{tab:main-result-acc}
% \input{tables/main_tab_acc}
% \end{table*}

In this way, we can
% transform the word probability difference into the sentence probability difference,
% The advantage 
% of 
% modeling the entire sentence's probability 
% which 
capture the semantics of the entire sentence, rather than only the pre-context information.
% prefix sequence.
The proof of Theorem~\ref{theo} is provided in Appendix \ref{pr:llm}.

\subsection{Time Complexity Analysis} 
\label{sec:compAna}
\name\ consists of two main stages: toxic word matching and filtering. Suppose we have an input text sequence $X$ of length $\vert X \vert$ and a toxic lexicon with $m$ toxic words.

In the {matching} stage, 
we enumerate substrings in $X$ and match them with toxic words in the lexicon,
whose largest length is $N_{max}$.
We enumerate substrings of length $\leq N_{max}$ for matching.
Considering that $N_{max} \ll \vert X \vert $, 
the number of enumerated substrings is linear to $\vert X \vert$.
For each substring, 
we only need to match it with the toxic word of the same length,
whose average number is assumed to be $\frac{m}{N_{max}}$.
Therefore,
the overall matching time complexity is $O(\frac{m}{N_{max}}\vert X \vert)$.

% The detection begins by enumerating the length of toxic words~(from $1$ to $7$). For each length $N$, it performs a linear $O(T)$ scan of string $X$ using a sliding window to generate candidate substrings. Each candidate is then checked against toxic words with the same length. 
% Therefore the matching step results in an overall $7\cdot O(T) \cdot S=S\cdot O(T)$ time complexity. 

In the {filtering} stage, we iteratively handle $M$ candidate \((w, l)\) pairs. In each iteration, for each pair \((w, l)\), we calculate the occurrence probability difference of the two words,
leading to a time complexity of $O(M\vert X \vert ^ 2)$ no matter whether BERT or LLMs are used.
The overall time complexity is thus $O(M^2\vert X \vert ^ 2)$.
Different from other LLM-based models that generate tokens with a time complexity of $O(\vert X \vert ^ 3)$, our method \name-LLM only uses LLMs to output the probability difference, which is highly more efficient.

% requiring $2M$ transformer forward passes ($O(T^2)$ each) to compute word/sentence probabilities. With at most $M$ iterations (removing one pair each time), the complexity is $2M^2\cdot O(T^2)$.

% In summary, the total time complexity is 
% Thus, the total complexity is $S\cdot O(T)+2M^2\cdot O(T^2)=O(T^2)$. While our algorithm's runtime is dominated by language model's forward passes, its logit-only design ensures markedly lower costs than generative approaches - a difference we measure experimentally.

\section{Experiment}
\label{sec:exp}
\subsection{Experimental Details}
\label{sec:exp_details}

\paragraph{Datasets}
We conduct experiments on \texttt{ToxicloakCN}~\cite{6} and \texttt{COLDataset}~\cite{14}.
For \texttt{ToxicloakCN}, the main results reported in this paper use the homophone-cloaked subset. Emoji-based perturbations are excluded from the main tables and discussed as a limitation.
% We use the keyword-augmented noisy part from \texttt{ToxicloakCN}~\cite{6} and \texttt{COLDataset}~\cite{14} as our test sets.
% Additionally,
Further,
for fairness,
we fine-tune the BERT-based CSC model on a subset of \texttt{CHSD}~\cite{CHSD} named as \texttt{CHSD-subset}.
Details on these datasets are given in Appendix~\ref{app:datasets}.
We additionally provide an out-of-domain and cross-resource support analysis in Appendix~\ref{app:ood}.

\paragraph{Baselines}
To our best knowledge, we are the first to unveil Chinese cloaked toxicity. 
Considering the relevance between our task and CSC task,
% Hence,
we thus compare our method with three categories of baselines.
% We compare our method with three categories of baselines.
(1) \textit{Naive}: Directly replace all candidate toxic words without any filtering.
(2) \textit{BERT-based method}: SCOPE~\cite{23} is a BERT-based CSC model that introduces an auxiliary task of Chinese pronunciation prediction to improve CSC task. We also finetune the pre-trained SCOPE model on \texttt{CHSD-subset}, obtaining the ft-SCOPE model.
(3) \textit{LLM-based method}: i) The prompt-based LLMs Baichuan2-7B-Base~\cite{yang2023baichuan}, Deepseek-V3~\cite{liu2024deepseek} and GPT-4o~\cite{hurst2024gpt}. ii) Simple-CSC~\cite{12} is a training-free and prompt-free LLM-based CSC model that considers character pronunciation and shape similarities.

\begin{table*}[t!]
\centering

\resizebox{0.85\linewidth}{!}{
\begin{tabular}{llrrrrrrrr}

\toprule
\multirow{3}{*}{Metric/\%} & \multirow{3}{*}{Method} & \multicolumn{4}{c}{\textbf{ToxicloakCN}} & \multicolumn{4}{c}{\textbf{COLDataset}} \\
% \hline
% \cline{3-10}
& & \multicolumn{2}{c}{Sentence} & \multicolumn{2}{c}{Character} & \multicolumn{2}{c}{Sentence} & \multicolumn{2}{c}{Character} \\
& 
& \small Det. & \small Cor.
& \small Det. & \small Cor.
& \small Det. & \small Cor.
& \small Det. & \small Cor.
% & DET & CORR & DET & CORR & DET & CORR 
\\
\midrule
\multirow{8}{*}{F1 score}
& Naive & 3.96&2.77&20.29&17.31&72.32&64.00&90.42&87.41\\
& SCOPE & 13.23&6.33&7.81&2.22&14.80&9.91&18.41&7.01 \\
& ft-SCOPE & 35.79&26.99&17.48&11.73&62.66&59.15&41.21&35.83 \\
& Simple-CSC & 47.81&41.28&50.18&43.77&59.60&57.36&71.60&69.28 \\
& Baichuan2-7B-Base & 2.62&2.18&3.46&0.77&1.26&1.03&5.31&0.90 \\
& Deepseek-V3&28.66&20.21&17.15&8.56&40.89&31.99&40.15&26.84 \\
& GPT-4o&	27.69&	15.98	&22.00&	9.66&	40.02&	29.49&	45.86&	29.92\\
% \hline
\cmidrule(lr){2-10}
& \name-BERT & \underline{62.65}&\underline{58.53}&\underline{69.77}&\underline{65.81}&\underline{87.06}&\underline{82.56}&\underline{91.00}&\underline{90.00}\\
& \name-LLM & \textbf{74.67}&\textbf{70.01}&\textbf{78.56}&\textbf{75.04}&\textbf{90.52}&\textbf{88.54}&\textbf{93.73}&\textbf{93.30} \\
\midrule
\multirow{8}{*}{Accuracy} 
& Naive & 7.28&6.41&81.08&80.69&71.87&64.00&98.73&98.38 \\
& SCOPE & 35.35&31.33&79.42&78.79&15.47&11.22&78.35&76.78 \\
& ft-SCOPE & 52.88&47.36&80.72&80.01&63.52&60.14&82.32&81.33 \\
& Simple-CSC & 66.38&63.58&98.13&97.97&53.85&51.99&97.15&96.97 \\
&Baichuan2-7B-Base&40.03&39.86&92.55&92.44&4.55&4.43&90.99&90.78\\
&Deepseek-V3&46.36&41.04&83.46&82.57&40.59&32.26&85.67&83.82 \\
& GPT-4o&	33.43&	25.01&	87.58&	86.55&	39.91&	29.74&	88.67&	86.71\\
% \hline
\cmidrule(lr){2-10}
& \name-BERT & \underline{76.32}&\underline{74.38}&\underline{98.66}&\underline{98.53}&\underline{84.65}&\underline{80.51}&\underline{98.94}&\underline{98.84} \\
& \name-LLM & \textbf{83.60}&\textbf{81.36}&\textbf{99.02}&\textbf{98.89}&\textbf{88.37}&\textbf{86.55}&\textbf{99.25}&\textbf{99.20} \\
\bottomrule
\end{tabular}
}

\caption{The main results of all baseline models and our method w.r.t. the F1 score/\% and the accuracy/\% metrics, where Det. and Cor. mean detection and correction. We highlight the \textbf{best} score in bold and underline the \underline{runner-up} one.}
\label{tab:cat_main}
\end{table*}

\paragraph{Setup}
We construct homophone graph by pypinyin, and utilize lexicons after manual correction and deduplication for toxicity matching.
Our filtering model consists of two versions: BERT-based and LLM-based. 
For the former,
we utilize bert-chinese-base~\cite{2}.
For the latter,
we use Baichuan2-7B-Base~\cite{yang2023baichuan} as our backbone model.
We provide a more detailed backbone comparison, including newer Qwen3.5 models, in Appendix~\ref{app:llm_backbones}.
% various LLMs, including LLaMA-3-8B-Instruct~\cite{llama3modelcard}, Qwen2.5-Instruct~(7B, 14B)~\cite{qwen2.5, qwen2} and Baichuan2-Base~(7B, 13B)~\cite{yang2023baichuan}. All models can be downloaded on Huggingface\footnote{\url{https://huggingface.co/}}. We finally choose Baichuan2-7B-Base model as backbone model of \name-LLM for the best performance.
None of the models' parameters have been modified, nor have any prompts been used, as our method is entirely training-free and prompt-free.
For SCOPE, we follow its original pre-training stage and finetune it on \texttt{CHSD-subset}.
For Simple-CSC, we employ it on Baichuan2-13B-Base as its original paper. All models can be downloaded on Huggingface\footnote{\url{https://huggingface.co/}}.
We also notice that there exist some other CSC baselines.
However,
some of them~\cite{21, 11} underperform the two used CSC baselines, while others~\cite{22, li2024c} lack complete pretraining datasets and source codes. Thus, we exclude them from comparison.
The prompt template used for prompt-based LLMs is presented in Appendix \ref{ap:prompt}. Our experiments are conducted on a server equipped with an Nvidia A800 80G GPU.

\subsection{Main Results}
We evaluate the model performance by F1 score and accuracy from two aspects (sentence-level and character-level), with each one containing both detection and correction tasks.
All reported F1 and accuracy values are micro-averaged over the full evaluation set, following the aggregated counts defined in Appendix~\ref{metrics}.
The sentence-level task requires the complete detection/correction of all Chinese cloaked characters in a sentence and keeps other non-cloaked characters not mis-detected/mis-corrected;
% The corrected sentence has to be equal to the target sentence; 
otherwise, the task fails.
For the character-level task, detection and correction are applied exclusively to individual characters. The effectiveness of the model is reflected by its ability to identify and correct more cloaked characters while minimizing modifications to non-toxic ones—the higher the true positive rate and the lower the false positive rate, the better the performance.
% \textbf{To be strict,
% the sentence-level task is more reliable than the character-level task.}
Details on the evaluation metrics can be found in Appendix \ref{metrics}.
% For the character-level task,
% we only perform detection/correction on single characters.
% The more cloaked characters 
% detected/corrected, the less non-toxic ones detected/corrected, the better the results. 
% \xc{Sentence-level measurement requires perfect detection/correction of all characters in a sentence, presenting a more rigorous test of model capability. In contrast, character-level evaluation assesses individual character's detection/correction, where the predominance of non-replaceable characters creates significant class imbalance.}
% For both metrics, the larger the values, the better the results.
% Given a sample pair of (source sentence, target sentence) and the corresponding corrected sentence, for the F1 score computation in the sentence level, if and only if the source sentence and the target sentence are identical, the sample is considered as negative.
% % instances, and vice versa. 
% The correction is true (including true positive and true positive) if and only if the corrected sentence is same as the target sentence. 
% This also holds for the character-level comparison, where the evaluation is performed on single Chinese characters. 
% And following this definition, the accuracy can be easily computed.
From Table \ref{tab:cat_main}, we observe:

(1) The Naive model 
generally performs poorly across the datasets, which shows the importance of candidate toxic words filtering.

(2) ft-SCOPE consistently outperforms SCOPE across all the metrics and datasets. 
This shows the significant domain gap between general CSC pretraining data and toxic speech corpora. 
\name-BERT, which is not fine-tuned, beats ft-SCOPE across all metrics.
% (also BERT-based and without any finetuning)
% outperform ft-SCOPE on \texttt{ToxicloakCN} and the sentence-level comparison on \texttt{COLDataset}.
% This is because 
% sentence-level detection and correction, but shows slightly lower effectiveness in character-level tasks.
Although fine-tuning helps memorize toxic knowledge,
it only memorizes 
fixed toxicity mapping patterns in the training data. However,
our model injected with homophone graph and toxic lexicon has stronger generalizability,
which can perceive more cloaked toxicities.

(3) 
% Similar to \name-LLM,
The LLM-based model Simple-CSC exhibits undesired results in some cases, which are even worse than BERT-based models.
For example, 
% w.r.t. the F1 score, 
in the sentence-level toxicity detection on the \texttt{COLDataset}, 
Simple-CSC achieves an accuracy of $53.85\%$ while that of ft-SCOPE is $63.52\%$.
This can be attributed to the limited cloaked toxic knowledge of LLMs, 
which is in accordance with the poor performance of 
prompt-based LLM models Baichuan2-7B-Base, Deepseek-V3 and GPT-4o.
% which has been pointed out in~\cite{zhao2024enhancing}.

% Simple-CSC is a training-free and prompt-free method that leverages phonetic similarity in Chinese characters, purely utilizing LLM's language modeling capability. However, its performance lags significantly behind ours, as our method incorporates toxic lexicon information during detection to precisely identify potential obfuscation locations, thereby enhancing effectiveness.

(4) 
\name-LLM and \name-BERT
% \name-LLM
take the first two ranks 
% leads other competitors 
in all contests.
On the one hand,
they use homophone graph and toxic lexicon to match all the potential toxic words in a given text sequence, which increases true positive rate.
On the other hand,
they further utilize a filtering step to mitigate the risk that transforms non-toxic words into toxic, which decreases false positive rate.
We further corroborate this ordering with a human evaluation on 200 sampled \texttt{ToxicloakCN} examples, which is reported in Appendix~\ref{app:human_eval}.

\begin{table*}[t!]
\centering
% \resizebox{0.8\linewidth}{!}{%
% \begin{tabular}{lcccccccc}
% \hline
% Dataset&\multicolumn{8}{c}{ToxicloakCN}\\\hline
% &\multicolumn{4}{c}{Detection}&\multicolumn{4}{c}{Correction}\\
% Metrics/\%&TP$\uparrow$&FP$\downarrow$&FN$\downarrow$&TN$\uparrow$&TP$\uparrow$&FP$\downarrow$&FN$\downarrow$&TN$\uparrow$\\\hline

% Naive&\textbf{2.71}&21.07&\textbf{0.21}&76.01&\textbf{2.27}&21.07&\textbf{0.65}&76.01\\
% \name-BERT&1.74&0.33&1.18&96.75&1.59&0.33&1.33&96.75\\
% \name-LLM&2.02&\textbf{0.20}&0.90&\textbf{96.88}&1.87&\textbf{0.20}&1.05&\textbf{96.88}\\
% \hline
% Dataset&\multicolumn{8}{c}{COLDataset}\\\hline
% &\multicolumn{4}{c}{Detection}&\multicolumn{4}{c}{Correction}\\
% Metrics/\%&TP$\uparrow$&FP$\downarrow$&FN$\downarrow$&TN$\uparrow$&TP$\uparrow$&FP$\downarrow$&FN$\downarrow$&TN$\uparrow$\\\hline

% Naive&\textbf{6.77}&1.06&\textbf{0.38}&91.79&\textbf{6.37}&1.06&\textbf{0.78}&91.79\\
% \name-BERT&6.04&0.09&1.11&92.76&5.92&0.09&1.23&92.76\\
% \name-LLM&6.32&\textbf{0.02}&0.83&\textbf{92.83}&6.27&\textbf{0.02}&0.88&\textbf{92.83}\\
% \hline
% \end{tabular}%
% }

\resizebox{\linewidth}{!}{%
\begin{tabular}{lcccccccccccc}
\hline
Dataset&\multicolumn{12}{c}{\textbf{ToxicloakCN}}\\\hline
&\multicolumn{6}{c}{Detection}&\multicolumn{6}{c}{Correction}\\
Metrics/\%&TP$\uparrow$&FP$\downarrow$&FN$\downarrow$&TN$\uparrow$&recall$\uparrow$&precision$\uparrow$&TP$\uparrow$&FP$\downarrow$&FN$\downarrow$&TN$\uparrow$&recall$\uparrow$&precision$\uparrow$\\\hline

Naive&\textbf{2.41}&18.74&\textbf{0.19}&78.67&\textbf{92.74}&11.39&\textbf{2.02}&18.74&\textbf{0.58}&78.67&\textbf{77.83}&9.74\\
\name-BERT&1.55&0.29&1.05&97.11&59.54&84.25&1.42&0.29&1.18&97.11&54.50&83.04\\
\name-LLM&1.79&\textbf{0.18}&0.80&\textbf{97.23}&69.09&\textbf{91.04}&1.67&\textbf{0.18}&0.93&\textbf{97.23}&64.13&\textbf{90.41}\\

\hline
Dataset&\multicolumn{12}{c}{\textbf{COLDataset}}\\\hline
&\multicolumn{6}{c}{Detection}&\multicolumn{6}{c}{Correction}\\
Metrics/\%&TP$\uparrow$&FP$\downarrow$&FN$\downarrow$&TN$\uparrow$&recall$\uparrow$&precision$\uparrow$&TP$\uparrow$&FP$\downarrow$&FN$\downarrow$&TN$\uparrow$&recall$\uparrow$&precision$\uparrow$\\\hline

Naive&\textbf{5.99}&0.94&\textbf{0.33}&92.74&\textbf{94.72}&86.50&\textbf{5.64}&0.94&\textbf{0.69}&92.74&\textbf{89.11}&85.77
\\
\name-BERT&5.35&0.08&0.98&93.59&86.83&99.51&5.24&0.08&1.09&93.59&85.51&99.50
\\
\name-LLM&5.6&\textbf{0.02}&0.73&\textbf{93.66}&88.42&\textbf{99.71}&5.55&\textbf{0.02}&0.78&\textbf{93.66}&87.66&\textbf{99.70}
\\
\hline
\end{tabular}%
}

\caption{Character-level analysis between the Naive method with \name\ on \texttt{ToxicloakCN} and \texttt{COLDataset}. T/F means True/False. P/N means Positive/Negative. All results are normalized by dividing the raw values by the total number of sentences and characters in the dataset. We highlight the \textbf{best} score in bold.}
\label{tab:cat-naive-fp}

\end{table*}

% \begin{table*}[t!]
% \centering
% \caption{Character-level analysis between the Naive method with \name\  on \texttt{COLDataset}.}
% \label{tab:naive-fp-c}
% \input{tables/naive_fp_COLD}
% \end{table*}

\subsection{Analysis of Filtering Stage}

As mentioned in Section \ref{sc:WordMatching}, 
the Naive method that includes toxic word matching only and discards candidate toxic word filtering could lead to small false negative rate at the cost of high false positive rate.
We next show the results of \textit{True/False Positive/Negative rate}, \textit{recall} and \textit{precision} in Table~\ref{tab:cat-naive-fp}.
Naive has better TP and FN rates than our methods,
which results in higher recall values because it regards all the candidate toxic substrings as the ``true'' toxic words.
However, this leads to over-detection and over-correction, which categorizes a large number of non-toxic words to be toxic.
Such indiscriminate filtering severely compromises the semantic integrity of benign sentences, making the Naive approach impractical for real-world scenarios where preserving the original meaning is crucial.
% Therefore, the precision of Naive is very poor.
Our methods further filter out candidate toxic words that are non-toxic and thus achieve better FP and TN rates, and higher precision scores.
For example, 
On the \texttt{ToxicloakCN} dataset,
in the correction task,
Naive has a precision of $9.74\%$, while that of our model \name-LLM is $90.41\%$, which is around 10$\times$ larger.
In summary,
our methods significantly improve precision, at the cost of a mild drop in recall,
which finally lead to larger F1 scores. 
Representative error cases for \name-BERT and \name-LLM are summarized in Appendix~\ref{app:case_study}.

% \begin{figure*}[t!]
%   \includegraphics[width=0.48\linewidth]{figs/T_WP_F1.pdf} \hfill
%   \includegraphics[width=0.48\linewidth]{figs/C_WP_F1.pdf}
%   \caption {Sentence level correction F1 score gap between \name-WP and \name\  on both datasets.}
%   \label{fig:pre}
% \end{figure*}

% \begin{figure*}[t!]
%   \includegraphics[width=0.48\linewidth]{figs/T_SR_F1.pdf} \hfill
%   \includegraphics[width=0.48\linewidth]{figs/C_SR_F1.pdf}
%   \caption {Sentence level correction F1 score gap between \name-SR and \name\  on both datasets.}
%   \label{fig:single-multi}
% \end{figure*}

\subsection{{Efficiency study}}
% We conduct theoretical analysis of \name's complexity in Section \ref{sec:compAna} and evaluate it here.

% lexicon大小 S 有害词长度 N candidate有害词个数 M 句子长度T

% \begin{figure*}[t!]
% \makebox[\textwidth][c]{\includegraphics[width=0.9\textwidth]{figs/ef_study.pdf}}
%   % \includegraphics[width=\linewidth]{figs/ab.pdf}
%   \caption {}
%   \label{fig:ef}
% \end{figure*}

\begin{figure}[t!]
  \centering
  \includegraphics[width=1.00\linewidth]{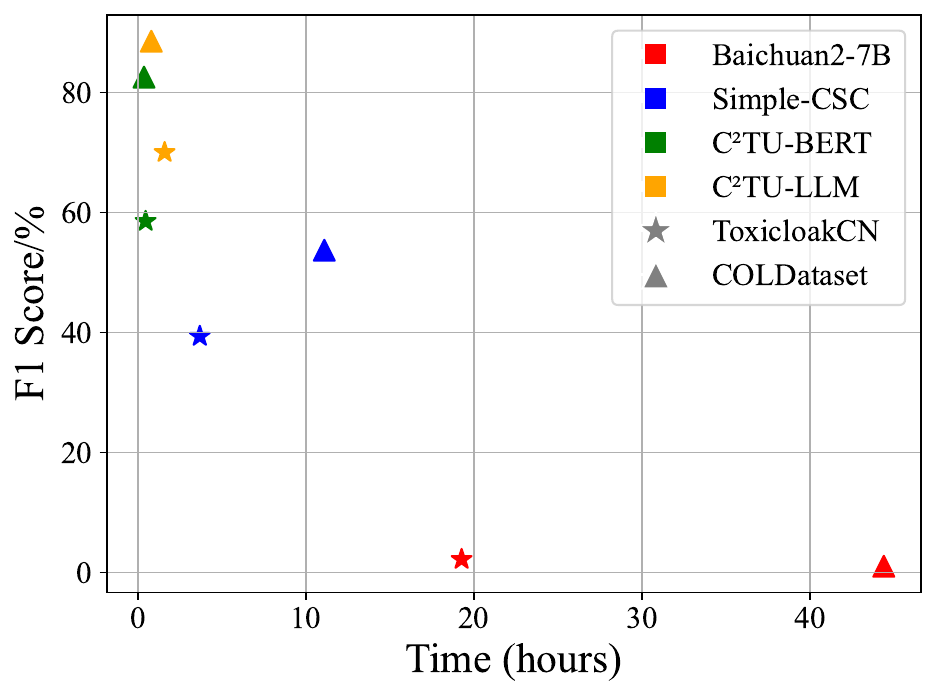}
  \vspace{-1em}
  \caption{Efficiency study on the sentence-level correction task. }
  \label{fig:efficiency}
  % \vspace{-0.5em}
\end{figure}

We next analyze the model efficiency.
The results are presented in Figure~\ref{fig:efficiency},
where $\blacktriangle$ indicates \texttt{COLDataset} and $\bigstar$ represents \texttt{ToxicloakCN}.
We use different colors to denote different methods including \name-LLM, \name-BERT, Simple-CSC and Baichuan2-7B-Base.
For fairness, all LLM-based methods are based on Baichuan2-7B-Base. 
From the figure, we observe:
(1) As LLM-based methods,
\name-LLM is significantly more efficient than Simple-CSC and Baichuan2-7B-Base.
This is because \name-LLM only uses LLM to output sentence probability difference,
while the other two methods perform token generation.
(2) \name-BERT is more efficient than \name-LLM due to the less number of model parameters, but it is at the cost of lower model effectiveness.
In summary, our models are highly effective and also efficient.

\subsection{Extension to Shape-Similar Cloaks}

To further test whether \name\ is tied to homophone-based cloaks, we run a supplementary experiment on shape-similar characters. We keep the matching and filtering pipeline unchanged and only replace the homophone graph with a shape-similarity graph extracted using the open-source SimilarCharacter\footnote{\url{https://github.com/contr4l/SimilarCharacter}} library. This setting isolates whether the framework can transfer to another relation graph without additional training.

\begin{table}[t!]
\centering
\resizebox{\columnwidth}{!}{%
\begin{tabular}{lcccc}
\toprule
Method & Sent-Det. & Sent-Cor. & Char-Det. & Char-Cor. \\
\midrule
Simple-CSC & 70.32 & 67.86 & 98.64 & 98.54 \\
C$^2$TU-BERT & \underline{89.97} & \underline{89.97} & \underline{99.42} & \underline{99.42} \\
C$^2$TU-LLM & \textbf{94.92} & \textbf{94.92} & \textbf{99.59} & \textbf{99.59} \\
\bottomrule
\end{tabular}%
}
\caption{Supplementary evaluation on \texttt{ToxicloakCN} with shape-similar cloaks. We extract the shape-similarity graph using SimilarCharacter and keep the rest of the framework unchanged.}
\label{tab:shape_similar}
\end{table}

The results in Table \ref{tab:shape_similar} show that \name\ still brings consistent gains under a different relation graph, indicating that the framework is not limited to homophone cloaks alone.

% \begin{figure*}[t!]
% \makebox[\textwidth][c]{\includegraphics[width=0.85\textwidth]{figs/ab_example.pdf}}
%  \caption {Sentence level correction accuracy gap between \name\ , \name-WP and \name-SR on both datasets. The BERT-based model is employed on bert-chinese-base and the LLM-based model is employed on Baichuan2-7B-Base.}
%   \vspace{-1em}
%  \label{fig:ab_sca}
% \end{figure*}

% \begin{figure*}[t]
% \makebox[\textwidth][c]{\includegraphics[width=0.9\textwidth]{figs/full_ab.pdf}}
%   \caption {Full results of ablation study.}
%   \label{fig:ab}
% \end{figure*}

\subsection{Ablation Study}

We also conduct an ablation study on \name\ to understand the characteristics of its main components.
Given the auto-regressive nature of LLMs, it is direct to predict the occurrence probability of a word based only on its preceding context $X_{\text{pre}}$. 
This helps us understand the importance of the sentence probability method that leverages full context in the filtering stage.
% of \name-LLM. 
We call this method \textbf{\name-WP} (\textbf{W}ord \textbf{P}robability). 
Further,
filtering candidate toxic words in Algorithm \ref{alg:fptw} involves multiple rounds. 
We introduce a single round strategy, in which each \((w, l)\) pair satisfying \(P_w<P_l\) is replaced. This helps us understand the importance of the iterative strategy for filtering. We call this method \textbf{\name-SR} (\textbf{S}ingle \textbf{R}ound). 

We then compare \name\ with \name-WP and \name-SR. 
Due to the space limitation, 
we only show the results on the accuracy of sentence-level correction in Figure \ref{fig:ab_sca}.
For other cases,
we observe similar results, detailed in Appendix~\ref{ap:full_ab}.
% which reflects the model's capability to correct entire sentences. 
% The reasons are that: (1) Correction is more challenging than detection. (2) Character-level scores tend to be inflated due to many unaltered characters between source and target sentences. (3) Compared to accuracy, the F1 score better reflects positive-class identification capability.
% The results are given . 
From the figure, we see that:
(1) 
\name-LLM significantly beats \name-WP.
This indicates that 
$X_{\text{tail}}$ also provides essential semantic information and 
leveraging the full text context for filtering candidate toxic words is necessary.
(2) For both \name-BERT and \name-LLM, they consistently outperform \name-SR across the datasets.
This is because \name\ replaces the \((w, l)\) pair with the largest probability difference in each round, 
which iteratively enhances the context of remaining \((w, l)\) pairs and the correction accuracy.

\begin{figure}[t!]
  \centering
  \includegraphics[width=1.00\linewidth]{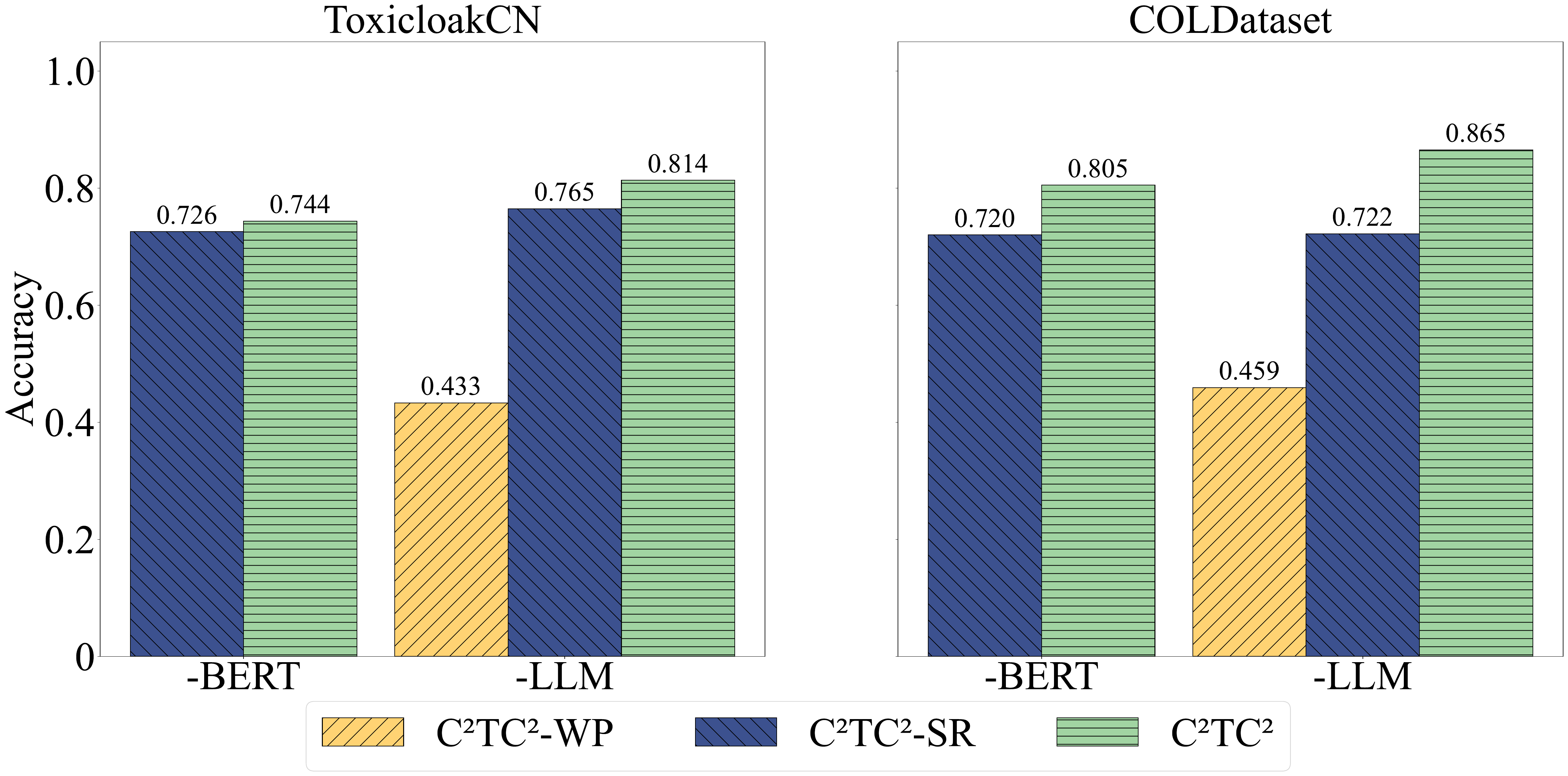}
  \vspace{-1em}
  \caption{Sentence level correction accuracy gap between \name\ , \name-WP and \name-SR on both datasets. The BERT-based model is employed on bert-chinese-base and the LLM-based model is employed on Baichuan2-7B-Base.}
  \label{fig:ab_sca}
  % \vspace{-0.5em}
\end{figure}

\section{Conclusion}
\label{sec:conc-lim}
In this paper, we presented \name, a training-free and prompt-free method for unveiling Chinese cloaked toxic contents on social platforms. Specifically, our method consists of two stages: Chinese cloaked toxicity matching and over-unveiling candidates filtering. 
For the former, we constructed the homophone graph to capture similar homophone relationships between Chinese characters. Subsequently, we matched toxic words via the homophone graph and toxic lexicon. 
To mitigate over-correction issues, we employed context-aware language models~(BERT-based and LLM-based) to filter candidate toxic words iteratively. 
Comprehensive experimental results show that \name\ achieves superior performance in Chinese cloaked toxicity unveiling task with high efficiency.

\section*{Limitations}
\label{ap:lim}
Our method is designed to restore cloaked variants of known toxic terms, rather than discover entirely new toxic expressions or rapidly evolving slang.
It mainly targets character-substitution cloaks, while other cloak types such as character-splitting, cross-lingual homophones, and emoji replacements remain outside the current framework and will be studied in future work.

\section*{Ethical Considerations}
\label{ap:impact}

Our original intent is to unveil cloaked toxic contents online in order to optimize cyber governance and purify the digital environment, given that cloaked toxic contents are indeed a pervasive phenomenon.

We acknowledge that the toxic word candidates identified in Stage 1 - derived from homophonic graph and a toxic lexicon - while comprehensive, inevitably include numerous false positives. This occurs because simple homophonic matching with toxicity inherently produces substantial misclassifications, which will result in a significant number of normal expressions being misclassified as toxicity.

To address this, we implement our Stage 2 filtering method to systematically eliminate normal expressions that were mistakenly flagged. Our experimental results in Table \ref{tab:cat-naive-fp} demonstrate that this filtering approach significantly reduces the false positive rate, albeit at some cost to true positive rate. This design choice prioritizes protecting users' rights to express legitimate content while maintaining detection efficacy.

From an academic research perspective, this represents the fullest extent of what we can achieve. Our technology is designed solely for academic research purposes, with no intended application in commercial or governmental contexts.

In the future, if this research ever be implemented in real-world content moderation systems, we maintain that human review remains indispensable. Our methodology can identify the vast majority of cloaked toxicity cases, which would then undergo human verification. This approach would ensure that every content flagged as toxic receives human validation.

To address the potential misuse of this technology for censorship, we require all users to obtain our authorization. We promise that we will not grant permission to government agencies or companies that are closely affiliated with governments.

% \clearpage

\section*{Acknowledgments}

This work was supported by Shanghai "Science and Technology Innovation Action Plan" Project under Grant No.23511100700.
\bibliography{custom}

\appendix

\section{Theoretical Proof}
\label{pr:llm}
\begin{proof}

Given a text sequence $X = [X_{\text{pre}},w, X_{\text{tail}}]$
and a toxic word $l$ with ${\mathrm{len}}(w) $= ${\mathrm{len}}(l)$ = $N$,
let $X'$ denote the new sequence with $w$ replaced by $l$. Then, the following equation holds:
\begin{equation}
\begin{aligned}
    {\mathrm{ProbDiff}}&(P_w,P_l) = \\
    &\hspace{0pt}\log P_{\mathrm{LLM}}(X)-\log P_{\mathrm{LLM}}(X')
\end{aligned}
\end{equation}
% \begin{equation}
% \begin{aligned}
%     {\mathrm{ProbDiff}}&(p_w,p_l)\\
%     =\ &\log P_{\mathrm{LLM}}(X)-\log P_{\mathrm{LLM}}(X')
% \end{aligned}
% \end{equation}
where 
$ P_{\mathrm{LLM}}(X)$ and $ P_{\mathrm{LLM}}(X')$
denote the probability of $X$ and $X'$ output by LLM, respectively. Both the probabilities can be calculated in the auto-regressive manner as Equation~\ref{eq:llmauto}.

Firstly, we derive the general formula for \(  P_{\mathrm{LLM}}(w | X_{\mathrm{pre}}, X_{\mathrm{tail}}) \):
\begin{equation}
\begin{aligned}
 P_{\mathrm{LLM}}(w|X_{\mathrm{pre}},X_{\mathrm{tail}}) &= \frac{ P_{\mathrm{LLM}}(X_{\mathrm{pre}},w,X_{\mathrm{tail}})}{ P_{\mathrm{LLM}}(X_{\mathrm{pre}},X_{\mathrm{tail}})}\\
&=\frac{ P_{\mathrm{LLM}}(X)}{ P_{\mathrm{LLM}}(X_{\mathrm{pre}},X_{\mathrm{tail}})}\\
\end{aligned}
\end{equation}

Analogously:
\begin{equation}
P_{\mathrm{LLM}}(l | X_{\mathrm{pre}}, X_{\mathrm{tail}})=\frac{ P_{\mathrm{LLM}}(X')}{ P_{\mathrm{LLM}}(X_{\mathrm{pre}},X_{\mathrm{tail}})}
\end{equation}

Therefore:
\begin{equation}
\begin{aligned}
&{\mathrm{ProbDiff}}(P_w,P_l)\\
=\ &\log P_w - \log P_l\\
=\ &\log P_{\mathrm{LLM}}(w|X_{\mathrm{pre}},X_{\mathrm{tail}})\ \\
&\hspace{66pt}- \log P_{\mathrm{LLM}}(l|X_{\mathrm{pre}},X_{\mathrm{tail}})\\
=\ &\log\frac{ P_{\mathrm{LLM}}(X)}{ P_{\mathrm{LLM}}(X_{\mathrm{pre}},X_{\mathrm{tail}})} \\
&\hspace{66pt}-\log\frac{ P_{\mathrm{LLM}}(X')}{ P_{\mathrm{LLM}}(X_{\mathrm{pre}},X_{\mathrm{tail}})}\\
%=\ &\log(\frac{ P_{\mathrm{LLM}}(X)}{ P_{\mathrm{LLM}}(X_{\mathrm{pre}},X_{\mathrm{tail}})}\cdot \frac{ P_{\mathrm{LLM}}(X_{\mathrm{pre}},X_{\mathrm{tail}})}{ P_{\mathrm{LLM}}(X')})\\
=\ &\log \frac{P_{\mathrm{LLM}}(X)}{P_{\mathrm{LLM}}(X')}\\
=\ &\log P_{\mathrm{LLM}}(X)-\log P_{\mathrm{LLM}}(X')\\
\end{aligned}
\end{equation}
Q.E.D.
\end{proof}

\section{Evaluation Metrics}
\label{metrics}
We evaluate the model performance by F1 score and accuracy from two aspects (sentence-level and character-level), with each one containing both detection and correction tasks, which is common metrics in the CSC field. Unless otherwise stated, all reported F1 and accuracy values are micro-averaged over the full evaluation set, with TP, FP, FN, and TN accumulated before computing the metrics.
%Firstly we define the computation method of T/F (true/false) P/N (positive/negative) in each level and task.

For sentence level, given the sample pair of (source sentence and target sentence) and the corresponding corrected sentence, if the length of corrected sentence is not equal to the target sentence, the result is FN for both detection and correction tasks. Otherwise, we have the following definition:

\paragraph{Sentence Level Detection}
(1) If the source sentence is different to the target sentence, the source sentence needs to be corrected. In this situation, if all wrong characters in source sentence is corrected (no matter correctly or not), and no proper character is over-corrected, the result is TP. Otherwise, the result is FN. (2) If the source sentence is identical to the target sentence, the source sentence doesn't need to be corrected. In this situation, if the corrected sentence is same to the target sentence, the result is TN. Otherwise, the result is FP.

\paragraph{Sentence Level Correction}
(1) If the source sentence is different to the target sentence, the source sentence needs to be corrected. In this situation, if all wrong characters in source sentence is corrected correctly, and no proper character is over-corrected, the result is TP. Otherwise, the result is FN. (2) If the source sentence is identical to the target sentence, the source sentence doesn't need to be corrected. In this situation, if the corrected sentence is same to the target sentence, the result is TN. Otherwise, the result is FP.

For character level, given the sample pair of (source character and target character) and the corresponding corrected character:

\paragraph{Character Level Detection}
(1) If the source character is different to the target character, the source character needs to be corrected. In this situation, if the corrected character is not same to the source character (no matter the correction is correct or not), the result is TP. Otherwise, the result is FN. (2) If the source character is identical to the target character, the source character doesn't need to be corrected. In this situation, if the corrected character is same to the target character, the result is TN. Otherwise, the result is FP.

\paragraph{Character Level Correction}
(1) If the source character is different to the target character, the source character needs to be corrected. In this situation, if the source character is corrected correctly, the result is TP. Otherwise, the result is FN. (2) If the source character is identical to the target character, the source character doesn't need to be corrected. In this situation, if the corrected character is same to the target character, the result is TN. Otherwise, the result is FP.

Based on these definitions, we can calculate the F1 score ($F_1$) and the accuracy ($Acc$) by:
\begin{equation}
    F_1=\frac{2pr}{p+r}
\end{equation}
\begin{equation}
    Acc=\frac{TP+TN}{TP+FN+FP+TN}
\end{equation}
where $p=\frac{TP}{TP+FP}$ and $r=\frac{TP}{TP+FN}$.

\section{Datasets}
\label{app:datasets}

We conduct experiments on two public datasets: \texttt{ToxicloakCN}~\cite{6} and \texttt{COLDataset}~\cite{14}, and finetune SCOPE~\cite{23} model on \texttt{CHSD-subset} constructed from \texttt{CHSD}~\cite{CHSD}. Details on these datasets are given in Table \ref{tab:dataset1}.

\begin{table}[htbp]
\centering
\begin{tabular}{ccc}
\hline
Dataset     & Size & Lexicon size  \\ \hline
\texttt{ToxicloakCN} & 4,586      & 387          \\
\texttt{COLDataset}  & 10,415     & 115          \\
\texttt{CHSD-subset} & 2,790      & 602          \\ \hline
\end{tabular}
\caption{Dataset size and lexicon size of \texttt{ToxicloakCN}, \texttt{COLDataset} and \texttt{CHSD-subset}.}
\label{tab:dataset1}
\end{table}

\paragraph{Homo-Graph}

We use pypinyin to capture tone-ignored pinyin of all Chinese characters and connect characters with identical or phonetically similar pinyins, which is defined in Section~\ref{sc:homo}. The details of homophone graphs are shown in Table \ref{tab:dataset2}.

\begin{table}[htbp]
\centering
\begin{tabular}{ccc}
\hline
Dataset     & Node size & Edge size  \\ \hline
\texttt{ToxicloakCN} & 3,613      & 141,683          \\
\texttt{COLDataset}  & 3,879     & 163,969          \\ \hline
\end{tabular}
\caption{Homo-graph size of \texttt{ToxicloakCN} and \texttt{COLDataset}.}
\label{tab:dataset2}
\end{table}

\paragraph{Toxic Lexicon}

(1) \texttt{ToxicloakCN} is an enhanced dataset derived from \texttt{ToxiCN}~\cite{15}, where JioNLP~\cite{jionlp} and NMSL\footnote{\url{https://github.com/THUzhangga/NMSL}} are applied to perform homophone substitution and semantically similar emoji substitution on toxic speech, respectively~\cite{6}. For the experiments in this paper, we use the homophone-substitution subset only, and exclude emoji-based perturbations from the reported main results.
For keyword-based substitution,
the toxic lexicon from \texttt{ToxiCN} is used as keywords.
For cleaner lexicon without cloak, we further correct and deduplicate the \texttt{ToxiCN}'s lexicon.
% For the \texttt{ToxicloakCN}~\cite{6} dataset, 
% In particular, we use the toxic lexicon collected by \texttt{ToxiCN}~\cite{15} with correction and deduplication processes. 
% \texttt{ToxiCN} categorizes toxic words into five types: general, LGBT, racism, region, and sexism. 
We first manually correct all toxic words in the original lexicon back to their proto-toxic words, and then retain each proto-toxic word.
(2) \texttt{COLDataset} is collected through web crawling based on keywords, comprising 37,480 contents and spanning a wide range of topics related to race, gender, and regional issues. 
Since no existing work has compiled a toxic lexicon for \texttt{COLDataset}, 
we instead utilize the crawler keywords used during dataset construction\cite{14} as toxic lexicon.
Based on the lexicon, 
we merge the validation set with the test set and apply JioNLP~\cite{jionlp} to perform homophone substitution and finally get the cloaked dataset, following the noise injection procedure of \texttt{ToxicloakCN}~\cite{6}.

% The final toxic lexicons are summarized in Appendix~\ref{sec:toxiclexicon}.

\begin{table}[H]
\centering
\begin{tabular}{ccc}
\hline
        & w/ T.L & w/o T.L \\ \hline
w/ C.L  & 1,706      & 542       \\
w/o C.L & 542      & 0       \\ \hline
\end{tabular}
\caption{The distribution of the \texttt{CHSD-subset}. T.L means \texttt{ToxiCN}'s lexicon, while C.L means \texttt{COLDataset}'s lexicon.}
\label{tab:chsd}
\end{table}

\paragraph{Dataset for Finetuning}

To ensure fair comparison between our model
and CSC model, 
we further fine-tune BERT-based CSC model SCOPE~\cite{23} on a subset of \texttt{CHSD}~\cite{CHSD},
which
% \texttt{CHSD} 
integrates datasets \texttt{COLDataset}~\cite{14}, \texttt{SWSR}~\cite{jiang2022swsr} and \texttt{CDIAL-BIASDATASET}~\cite{CDIAL} to balance the distribution difference of toxic contents (see Table~\ref{tab:chsd}). 
To construct the \texttt{CHSD-subset}, 
we first filter out samples that appear in the part of \texttt{COLDataset} used in the test set. 
Next, we select samples based on toxic keywords from the two lexicons of \texttt{ToxicloakCN} and \texttt{COLDataset}, ensuring that our fine-tuning data contains relevant toxic content. 
Then we merge the two lexicons 
% of \texttt{ToxicloakCN} and \texttt{COLDataset} 
and apply JioNLP~\cite{jionlp} to introduce noise. 
Note that we use the original lexicon of \texttt{ToxiCN} here. 
Finally, we construct \texttt{CHSD-subset} with 2790 samples that consists of 80\% training and 20\% validation data. Then we finetune the pre-trained SCOPE model with following settings: learning rate=$5e-5$, batch size=$32$, accumulate grad batches=$2$, epoch=$20$, warmup proporation=$0.1$. We first train the SCOPE model and it achieves a character-level detection F1-score of 0.8007 and a character-level corection F1-score of 0.7810 on CSC dataset. We then conducted fine-tuning on CHSD-subset and it achieves a character-level detection F1-score of 0.6026 and a character-level correction F1-score of 0.5531. 
% then split it into training set and validation set using an 8:2 ratio.
% \end{itemize}
Details of \texttt{CHSD-subset} are presented in Table \ref{tab:chsd}.

\section{Out-of-Domain and Cross-Resource Analysis}
\label{app:ood}

To provide additional support for the robustness of our matching-and-filtering framework, we examine how much overlap exists between the homophone graphs built from \texttt{COLDataset} and \texttt{ToxiCN}, as summarized in Table~\ref{tab:graph_overlap}. Although the two datasets differ, their commonly used character sets overlap substantially.

\begin{table}[htbp]
\centering
\begin{tabular}{lr}
\toprule
Quantity & Value \\
\midrule
\texttt{COLDataset} graph nodes & 3,879 \\
\texttt{ToxiCN} graph nodes & 3,613 \\
Intersection & 3,118 \\
Union & 4,374 \\
\texttt{COLDataset}-only & 761 \\
\texttt{ToxiCN}-only & 495 \\
Jaccard similarity & 0.7128 \\
\bottomrule
\end{tabular}
\caption{Overlap between the homophone graphs constructed from \texttt{COLDataset} and \texttt{ToxiCN}.}
\label{tab:graph_overlap}
\end{table}

We further conduct a cross-resource experiment on \texttt{ToxicloakCN} using the \texttt{ToxiCN} lexicon, as shown in Table~\ref{tab:cross_resource}. We compare three homophone graphs: the original \texttt{ToxiCN} graph, a joint graph formed by taking the union of the \texttt{COLDataset} and \texttt{ToxiCN} graphs, and a general graph built from all CJK Unified Ideographs in U+4E00--U+9FFF using pypinyin. The general graph contains 20,992 nodes and 4,460,324 edges.

\begin{table}[t!]
\centering
\resizebox{\columnwidth}{!}{%
\begin{tabular}{lcccc}
\toprule
Method & Sent-Det. & Sent-Cor. & Char-Det. & Char-Cor. \\
\midrule
w/ \texttt{ToxiCN} graph & \textbf{74.67} & \textbf{70.01} & 78.56 & \textbf{75.04} \\
w/ Joint graph & 72.33 & 68.44 & \textbf{82.97} & 74.69 \\
w/ General graph & 72.33 & 68.44 & \textbf{82.97} & 74.69 \\
\bottomrule
\end{tabular}%
}
\caption{Cross-resource evaluation on \texttt{ToxicloakCN} using the \texttt{ToxiCN} lexicon. We report F1 scores (\%).}
\label{tab:cross_resource}
\end{table}

The results suggest that a broader graph does not necessarily improve performance, since the contextual filter can already suppress many rare or implausible candidates.
More broadly, Chinese characters and their pronunciations form a finite and relatively stable inventory, so temporal drift is more likely to affect the toxic lexicon than the homophone relation itself.

\section{Representative Error Cases}
\label{app:case_study}

We manually inspect aligned original-cloaked sentence pairs and summarize representative errors from \name-BERT and \name-LLM in Table~\ref{tab:case_study}.

\begin{table*}[t!]
\centering
\scriptsize
\setlength{\tabcolsep}{3pt}
\begin{tabularx}{\textwidth}{l >{\raggedright\arraybackslash}p{2.4cm} >{\raggedright\arraybackslash}X >{\raggedright\arraybackslash}X >{\raggedright\arraybackslash}X >{\raggedright\arraybackslash}X}
\toprule
Dataset & Error type & Original Sentence & Cloaked Sentence & \name-BERT & \name-LLM \\
\midrule
\texttt{ToxicloakCN} & Short-span phonetic confusion &
连说说也不准？黑吹好猖獗 &
连说说也不准？嗨吹好猖獗 &
连说说也不准？嗨吹好猖獗 &
连说说也不准？黑吹好猖獗 \\
\texttt{COLDataset} & Polyphonic ambiguity &
冷暴力是不好，但是对男对女都是一样的啊，很多人强调用冷暴力对女生这个点，不理解 &
冷暴力是不好，但是对男对女都是一样的啊，很多人强调用冷暴力对女省这个点，不理解 &
冷暴力是不好，但是对男对女都是一样的啊，很多人强调用冷暴力对女性这个点，不理解 &
冷暴力是不好，但是对男对女都是一样的啊，很多人强调用冷暴力对女性这个点，不理解 \\
\texttt{ToxicloakCN} & Dialectal overgeneralization &
你这个喃蛮编的故事挺好的 &
你这个喃满编的故事挺好的 &
你这个南蛮编的故事挺好的 &
你这个南蛮编的故事挺好的 \\
\texttt{ToxicloakCN} & Overlapping candidates &
凭什么你觉得是女拳师你就用娘们侮辱人？[捂嘴]我还觉得你是男拳师呢，哪个词伺候？[捂嘴] &
凭什么你觉得是女权师你就用娘们侮辱人？[捂嘴]我还觉得你是南犬师呢，哪个词伺候？[捂嘴] &
凭什么你觉得是女权师你就用娘们侮辱人？[捂嘴]我还觉得你是男权师呢，哪个词伺候？[捂嘴] &
凭什么你觉得是女权师你就用娘们侮辱人？[捂嘴]我还觉得你是男权师呢，哪个词伺候？[捂嘴] \\
\bottomrule
\end{tabularx}

\caption{Representative error cases of \name-BERT (\texttt{bert-chinese-base}) and \name-LLM (\texttt{Baichuan2-7B-Base}).}
\label{tab:case_study}
\end{table*}

These examples suggest that the remaining failures are mostly caused by phonetic ambiguity, dialectal overgeneralization, and overlapping candidate spans.

\section{Prompt Template}
\label{ap:prompt}
Our prompt-based framework utilizes the following template structure (optimized version, generated by GPT-4o) to unveiling the cloaked toxicity:

\begin{table}[H]
\centering
\footnotesize
\phantomsection

\begin{tcolorbox}[colback=white!95!gray,colframe=gray!50!black,rounded corners, title={Prompt template for prompt-based method.
}]
% \begin{lstlisting}[breaklines=true, xleftmargin=0pt, breakindent=0pt, columns=fullflexible, mathescape]
- System: \\

任务：

针对中文敏感信息中带掩盖的错别字纠正。

要求：

1. 需要严格保证输入句子和输出句子长度一致；

2. 只需要返回纠正后的句子，不要输出任何其他内容。\\

- User: <Input sentence>

- Assistant: <Output sentence>
% \end{lstlisting}
\end{tcolorbox}
\caption{Prompt template for prompt-based method.
}

\end{table}

\section{Pseudocode of Matching and Filtering}
\subsection{Toxic Words Matching Algorithm}

\begin{algorithm}[H]
\caption{Toxic Words Matching}
\label{alg:stwma}

\renewcommand{\algorithmicrequire}{\textbf{Input:}}
\renewcommand{\algorithmicensure}{\textbf{Output:}}

\begin{algorithmic}[1]

\REQUIRE User comment $X$, lexicon $\mathcal L$, homophone graph $\mathcal G$
\ENSURE Potentially cloaked toxic words $W_p$
\STATE $W \gets {\rm{SlidingWindow}}(X), W_p \gets \emptyset$

\FOR{each $w^{(i)}$ in $W$, each $l^{(j)}$ in $\mathcal L$}
        \IF{${\rm{len}}(w^{(i)})={\rm{len}}(l^{(j)})$}
            \STATE $N\gets {\rm{len}}(w^{(i)}), flag\gets {\rm True}$
            \FOR{each $k \in \{1,2,\cdots,N\}$}
                \IF{$\mathcal G.{\rm{HasEdge}}(w^{(i)}_k, l^{(j)}_k)\ne 1$}
                    \STATE $flag\gets {\rm False}$
                \ENDIF
            \ENDFOR
            \IF{$flag = {\rm True}$}
                \STATE $W_p \gets W_p \cup\{(w^{(i)}, l^{(j)})\}$ 
            \ENDIF
        \ENDIF
\ENDFOR
\RETURN $W_p$

\end{algorithmic}

\end{algorithm}

\subsection{Filtering Toxic Words Algorithm}

\begin{algorithm}[H]
\caption{Filtering Toxic Words}
\label{alg:fptw}

\renewcommand{\algorithmicrequire}{\textbf{Input:}}
\renewcommand{\algorithmicensure}{\textbf{Output:}}

\begin{algorithmic}[1]

\REQUIRE $X, W_p,  P(\cdot)$
\ENSURE Corrected text $X'$
\STATE Initialize $X' \gets X$
\WHILE{$W_p$ is not empty}
    \FOR{each tuple $(w^{(i)}, l^{(j)})$ in $W_p$}
        \STATE $X_{\text{pre}}, X_{\text{tail}}\gets \text{GetPreTail}(X',w^{(i)})$
        \STATE $P_{w^{(i)}} \gets  P(w^{(i)} \mid X_{\text{pre}}, X_{\text{tail}})$
        \STATE $P_{l^{(i)}} \gets  P(l^{(i)} \mid X_{\text{pre}}, X_{\text{tail}})$
        \STATE $d_i \gets \text{ProbDiff}(P_{w^{(i)}}, P_{l^{(i)}})$
    \ENDFOR
    \STATE Get the index with the most significant difference: $k\gets \mathop{\text{argmin}}\limits_{i} \ d_i$
    \IF{$P_{w^{(k)}} < P_{l^{(k)}}$}
        \STATE $X_{\text{pre}}, X_{\text{tail}}\gets \text{GetPreTail}(X',w^{(k)})$
        \STATE Replace: $X' \gets [X_{\text{pre}}, l^{(k)}, X_{\text{tail}}]$
        \STATE $W_p\gets W_p \setminus \{(w^{(k)},l^{(k)})\}$
    \ELSE
        \STATE break
    \ENDIF
\ENDWHILE
\RETURN $X'$
\end{algorithmic}
\end{algorithm}

\begin{table*}[t!]
\centering

\resizebox{\linewidth}{!}{
\begin{tabular}{lllrrrrrrrr}

\toprule
\multirow{3}{*}{Metric/\%} & \multirow{3}{*}{Method}  & \multirow{3}{*}{Model} & \multicolumn{4}{c}{\textbf{ToxicloakCN}} & \multicolumn{4}{c}{\textbf{COLDataset}} \\
% \hline
% \cline{3-10}
& & & \multicolumn{2}{c}{Sentence} & \multicolumn{2}{c}{Character} & \multicolumn{2}{c}{Sentence} & \multicolumn{2}{c}{Character} \\
&& 
& \small Det. & \small Cor.
& \small Det. & \small Cor.
& \small Det. & \small Cor.
& \small Det. & \small Cor.
% & DET & CORR & DET & CORR & DET & CORR 
\\
\midrule

% Dataset&&\multicolumn{4}{c}{ToxicloakCN}&\multicolumn{4}{c}{COLDataset}\\\hline
% &&\multicolumn{2}{c}{Sentence}&\multicolumn{2}{c}{Character}&\multicolumn{2}{c}{Sentence}&\multicolumn{2}{c}{Character}\\
% Metrics/\%&&D-F&C-F&D-F&C-F&D-F&C-F&D-F&C-F\\\hline

\multirow{5}{*}{F1 score}&Simple-CSC& Baichuan2-13B & 47.81&41.28&50.18&43.77&59.60&57.36&71.60&69.28 \\
\cmidrule(lr){2-11}
&Prompt-based& Deepseek-V3&28.66&20.21&17.15&8.56&40.89&31.99&40.15&26.84 \\
\cmidrule(lr){2-11}
&\multirow{3}{*}{\name-LLM}&LLaMA3-8B&68.74&64.30&74.44&70.98&89.55&86.25&92.74&91.97\\

&&Qwen2.5-7B&67.20&63.52&73.46&70.19&89.37&86.30&92.41&91.73\\

&&Baichuan2-7B&\textbf{74.67}&\textbf{70.01}&\textbf{78.56}&\textbf{75.04}&\textbf{90.52}&\textbf{88.54}&\textbf{93.73}&\textbf{93.30} \\
\midrule

\multirow{5}{*}{Accuracy}&Simple-CSC&Baichuan2-13B& 66.38&63.58&98.13&97.97&53.85&51.99&97.15&96.97 \\
\cmidrule(lr){2-11}
&Prompt-based&Deepseek-V3&46.36&41.04&83.46&82.57&40.59&32.26&85.67&83.82 \\
\cmidrule(lr){2-11}
&\multirow{3}{*}{\name-LLM}&LLaMA3-8B&79.83&77.71&98.85&98.73&87.02&83.98&99.14&99.06\\

&&Qwen2.5-7B&78.81&77.04&98.80&98.69&86.94&84.12&99.10&99.03\\

&&Baichuan2-7B&\textbf{83.60}&\textbf{81.36}&\textbf{99.02}&\textbf{98.89}&\textbf{88.37}&\textbf{86.55}&\textbf{99.25}&\textbf{99.20}\\

\hline
\end{tabular}%
}

\caption{The results of \name-LLM with different LLMs w.r.t. F1 score and accuracy metrics, where Det. and Cor. mean detection and correction. We highlight the \textbf{best} score in bold.}
\label{tab:cat_llms}
\end{table*}

\section{Performance Across LLMs}
\label{app:llm_backbones}

We evaluate the performance of \name-LLM across different LLMs, including LLaMA3-8B~\cite{llama3modelcard}, Qwen2.5-7B~\cite{qwen2.5} and Baichuan2-7B~\cite{yang2023baichuan}.
We take Simple-CSC and Deepseek-V3 as our baselines, as they are also LLM-based. The results are shown in Table \ref{tab:cat_llms}.
While Simple-CSC employs Baichuan2-13B-Base and Deepseek-V3 is 671B,
our method consistently outperforms them with different LLMs of smaller parameter scales (7B and 8B).
This further shows the effectiveness of our method with the homophone graph and Chinese toxic lexicon, which inject knowledge of cloaked toxicity into LLMs.
We also report Qwen3.5 backbones~\cite{qwen3} on \texttt{ToxicloakCN} in Table~\ref{tab:qwen35_toxicloakcn}.

\begin{table}[t!]
\centering
\resizebox{\linewidth}{!}{
\begin{tabular}{lrrrr}
\toprule
Model & Sent-Det. & Sent-Cor. & Char-Det. & Char-Cor. \\
\midrule
Baichuan2-7B & \textbf{74.67} & \textbf{70.01} & 78.56 & \textbf{75.04} \\
Qwen3.5-4B & 69.31 & 64.83 & 81.39 & 71.53 \\
Qwen3.5-7B & 71.01 & 66.65 & \textbf{81.82} & 72.52 \\
\bottomrule
\end{tabular}%
}

\caption{Recent Qwen3.5 backbone comparison on \texttt{ToxicloakCN}. We report F1 scores (\%).}
\label{tab:qwen35_toxicloakcn}
\end{table}

The newer Qwen3.5 models still lag behind \name-LLM with Baichuan2-7B on this benchmark, which suggests that the task benefit comes more from the matching-and-filtering design than from a newer general-purpose backbone alone.

\section{Human Evaluation}
\label{app:human_eval}

Given that our evaluation metric for unveiling is based on rule matching rather than semantics, some sentences may not be corrected to their original form even though the unveiled outputs are valid. Therefore, we conducted a human evaluation based on sampling.
Specifically, we randomly select 200 samples from the \texttt{ToxicloakCN} dataset and perform unveiling using three models: Simple-CSC, \name-BERT, and \name-LLM, which are the three strongest systems under automatic evaluation.

For evaluation, we recruit three evaluators (two female and one male) with prior knowledge of Chinese toxicity. All of them are highly interested in cloaked toxicity topic and volunteer to undergo the experiment. They are recruited as Research Assistants (RAs) and remunerated in accordance with the wage standards prescribed by local laws. 

All RAs receive detailed task instructions and are shown the toxic lexicon. During assessment, each evaluator is presented with five versions of each sentence: the raw sentence, cloaked sentence, and the three model-generated unveiled sentences. They are then asked to label each model's output: marking it as "1" if they believed the sentence contains no remaining cloaking, and "0" otherwise. The final evaluation result for each sentence is determined by majority voting among the three evaluators' labels.
Table~\ref{tab:human-eval} reports the final human evaluation results, while Table~\ref{tab:human-eval-agreement} reports inter-annotator agreement statistics for the same sample.
The larger sample size reduces variance, and the same ranking remains: \name-LLM performs best, followed by \name-BERT and then Simple-CSC.

\begin{table}[t!]
\centering
\resizebox{1\linewidth}{!}{
\begin{tabular}{lcccc}
\toprule
Model  & Correct & Incorrect & Accuracy (\%) & 95\% CI (\%) \\
\midrule
Simple-CSC  & 154 & 46 & 77.00 & [70.69, 82.29] \\
\name-BERT  & 168 & 32 & 84.00 & [78.29, 88.43] \\
\name-LLM  & 178 & 22 & 89.00 & [83.91, 92.62] \\
\bottomrule
\end{tabular}
}
\caption{Human evaluation on 200 sampled examples from \texttt{ToxicloakCN}.}
\label{tab:human-eval}
\end{table}

\begin{table}[htbp]
\centering
\resizebox{\columnwidth}{!}{
\begin{tabular}{lccc}
\toprule
Model & Fleiss' Kappa & Observed Agreement $\bar{P}$ & Chance Agreement $P_e$ \\
\midrule
\name-LLM & 0.87 & 0.97 & 0.79 \\
\name-BERT & 0.88 & 0.97 & 0.72 \\
Simple-CSC & 0.94 & 0.98 & 0.64 \\
\bottomrule
\end{tabular}
}
\caption{Inter-annotator agreement for human evaluation.}
\label{tab:human-eval-agreement}
\end{table}

\begin{figure*}[t!]
\makebox[\textwidth][c]{\includegraphics[width=1\textwidth]{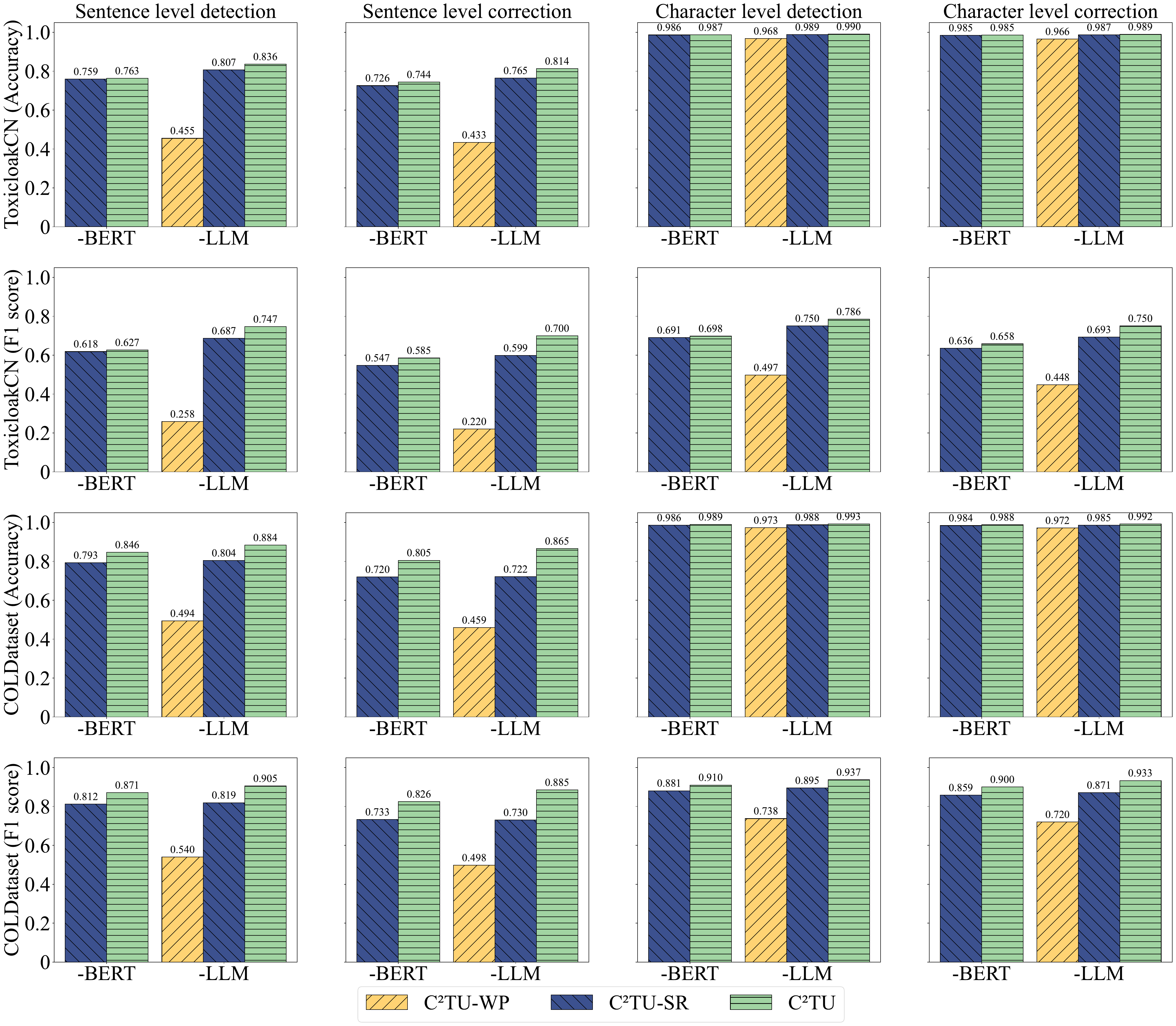}}
  \caption {Full results of ablation study.}
  \label{fig:full_ab}
\end{figure*}

\section{Full Results of Ablation Study}
\label{ap:full_ab}

Figure~\ref{fig:full_ab} summarizes the full ablation study between \name, \name-WP and \name-SR on both datasets. The \name-BERT is employed on bert-chinese-base and the \name-LLM is employed on Baichuan2-7B-Base.

% \section{Toxic Lexicon}
% \label{sec:toxiclexicon}

% We present toxic lexicons of \texttt{ToxicloakCN} and \texttt{COLDataset} here, splitting each lexicon by toxic words' length. Specifically, the toxic lexicon of \texttt{ToxicloakCN} is derived from the lexicon proposed in \texttt{ToxiCN}, with additional noise reduction and deduplication. Each toxic word is restored to its proto form, as the original lexicon contains a large number of cloaked toxic words. The toxic lexicon of \texttt{COLDataset}, on the other hand, is constructed from the crawler keywords that were used during the data collection process.

% \label{app:lex:cold}
%  \input{tables/COLD_lexicon}
% %\subsection{\texttt{ToxicloakCN}'s Lexicon}
% \label{app:lex:cloakcn}
% \input{tables/ToxiclaokCN_lexicon}

% %\subsection{\texttt{COLDataset}'s Lexicon}

\end{CJK*}
\end{document}